\documentclass[11pt]{article}

\usepackage[final]{acl}

\usepackage{times}
\usepackage{latexsym}
\usepackage{amssymb}
\usepackage{amsmath}
\usepackage{breqn}
\usepackage{mdframed}

\usepackage[T1]{fontenc}

\usepackage[utf8]{inputenc}

\usepackage{microtype}

\usepackage{inconsolata}

\usepackage[ruled,vlined]{algorithm2e}
\SetKwInOut{Parameter}{Hyper parameters}
\usepackage{graphicx}
\usepackage{booktabs}
\usepackage{xspace}
\usepackage{arydshln}

\usepackage{amssymb}
\usepackage{pifont}
\newcommand{\cmark}{\ding{51}}%
\newcommand{\xmark}{\ding{55}}%

\newcommand{\met}{\textsc{EffComp}\xspace}

\title{\met: Efficient Prompt Compression via Hybrid Reinforcement \& Supervised Learning}

\author{
  \textbf{Romtham Tungsununtham$^{1}$,
  Piyawat Lertvittayakumjorn$^{2}$,} \\
  \textbf{Atiwong Suchato$^{1}$,
  Peerapon Vateekul$^{1}$\thanks{Corresponding author}} \\
  $^{1}$ Department of Computer Engineering, Faculty of Engineering, \\
  Chulalongkorn University, Thailand \\
  $^{2}$ Google, United States \\
  \texttt{6672061921@student.chula.ac.th, piyawat@google.com,} \\
  \texttt{atiwong.s@chula.ac.th, peerapon.v@chula.ac.th}
}

\begin{document}
\maketitle
\begin{abstract}
Prompt compression is aimed at reducing input prompt lengths to enable cheaper, faster, and potentially higher-quality LLM predictions.
However, existing prompt compression methods are often limited by modest compression gains, a risk of hallucination, and/or high compression latency. 
We propose \met, an \textbf{Eff}icient prompt \textbf{Comp}ression framework for RAG-based question answering (QA).
\met employs BERT-style document reranker and sentence selector models to allow fast extractive prompt compression at the sentence level.
Its extractive nature prevents hallucinations in the compressed prompts. Additionally, the training process using a hybrid reinforcement and supervised learning approach is designed to optimize the compression ratio while preserving LLM accuracy. 
Experiments on various QA datasets demonstrate that \met outperforms state-of-the-art prompt compression methods achieving the best trade-off between compression ratio and accuracy with minimal latency. This makes it practical for real-world applications.
\end{abstract}

\section{Introduction}
Retrieval-augmented generation (RAG) enhances large language models (LLMs) by grounding their predictions in retrieved external documents to reduce hallucinations and address domain gaps~\citep{NEURIPS2020_6b493230,pmlr-v119-guu20a,izacard-grave-2021-leveraging}. However, the retrieved documents could often be long, redundant, and partially irrelevant~\citep{barnett2024seven,lertvittayakumjorn-etal-2025-towards}. The resulting lengthy augmented prompts not only undermine the quality of LLM predictions~\citep{shi2023large} but also increase computation costs and inference latency~\citep{keles2023selfattention}.
Therefore, recent research has proposed prompt compression methods to reduce the input prompt lengths (e.g., by strategically compressing the retrieved contexts) to address these efficiency and quality problems~\citep{li-etal-2025-prompt,wingate-etal-2022-prompt,chevalier-etal-2023-adapting}.

Despite these advancements, existing methods face several challenges and trade-offs. 
Abstractive prompt compressors generate a shorter context from a given long one, but they risk fabricating details or answering the questions themselves~\citep{xu2024recomp,alansari2025largelanguagemodelshallucination}.
Extractive prompt compressors, by contrast, preserve grounding by selecting and keeping the most important parts of the original context, but they often yield only modest compression ratios, e.g., 2x--7x in \cite{jiang-etal-2024-longllmlingua,pan-etal-2024-llmlingua,fei2025efficient,zhao-etal-2025-dac,Zhao_Wu_Xu_2025}.  
Furthermore, some compressors rely on large models or iterative algorithms, which can cause substantial latency and offset the benefits of compression~\citep{yoon-etal-2024-compact}.

To overcome these challenges, we propose \met, a novel prompt compression framework for RAG-based question answering (QA). The framework aims to boost the efficiency of prompt compression and subsequent LLM inference while maintaining (or even improving) the quality of the final output.
\met consists of two main steps. The first step reranks and filters the retrieved documents using a reranker. Then the second step selects important sentences in the remaining documents and concatenates them to form a compressed context for LLM inference.
The novelty of \met also lies in the training process of its sentence selector. This process first intermediately finetunes the model on a QA dataset and finally finetunes it using a hybrid reinforcement and supervised learning approach.
We compare our framework against reranker-only and other state-of-the-art prompt compression methods on four QA datasets. The experiments show that \met achieves higher QA accuracy and a strong compression ratio, outperforming other compression methods.
Overall, our contributions are threefold:

We introduce \met, a sentence-level extractive prompt compression framework for RAG-based QA. Its key innovations include (1) a sentence selector that jointly considers all shortlisted documents, (2) a reward function tailored to prompt compression for RAG-based QA, (3) a training data construction strategy for effective compressor learning, and (4) a hybrid RL-SFT training recipe that combines supervised and reward-driven learning.
(See Section~\ref{sec:method})

We evaluate \met on four QA datasets,
demonstrating that it achieves substantially high extractive compression while maintaining strong QA performance. (See Section~\ref{sec:experiments})

We conduct additional analyses to test performance of \met on LongBench, its performance of \met with commercial target LLMs, the ability to capture relevant context sentences, and the effects of different training strategies of \met. (See Section~\ref{sec:analysis})

\section{Problem Formulation}
In a RAG-based QA setting, let $q$ denote a question (so-called an input query) and $D = \langle d_1, d_2, \dots, d_n \rangle$ denote a list of $n$ retrieved documents ordered by the retrieval scores decreasingly. The full context $\mathbb{D}$ before compression is the concatenation of all $d_i \in D$ and has the length of $|\mathbb{D}|$ tokens. RAG combines $q$ and $\mathbb{D}$ to form an input prompt and feeds it to a target LLM $M$, which then outputs $M(q, \mathbb{D})$ to answer the question $q$. 
Prompt compression aims to find a compressed context $\mathbb{C}$ such that $|\mathbb{C}|\;\mathord{<}\;|\mathbb{D}|$ and $M(q, \mathbb{C})$ is a correct answer to $q$. 
We measure correctness using accuracy, where the accuracy (Acc) for a given context $X$ equals 1 if $M(q, X)$ contains the reference answer of $q$ (denoted as $y_q$) after text normalization\footnote{Following the evaluation script for SQuAD version 2.0 \citep{rajpurkar-etal-2018-know}, we perform text normalization by lowercasing the text, removing punctuation and articles, and collapsing extra whitespace.};
otherwise, 0. Following \citet{jiang-etal-2024-longllmlingua}, we define the compression ratio as $|\mathbb{D}|/|\mathbb{C}| = 1/\tau$.
\met aims to optimize the correctness metric while achieving a competitively high compression ratio.

\section{\met: The Efficient Prompt Compression Framework}
\label{sec:method}
\begin{figure*}[t]
    \centering
    \includegraphics[width=0.93\linewidth]{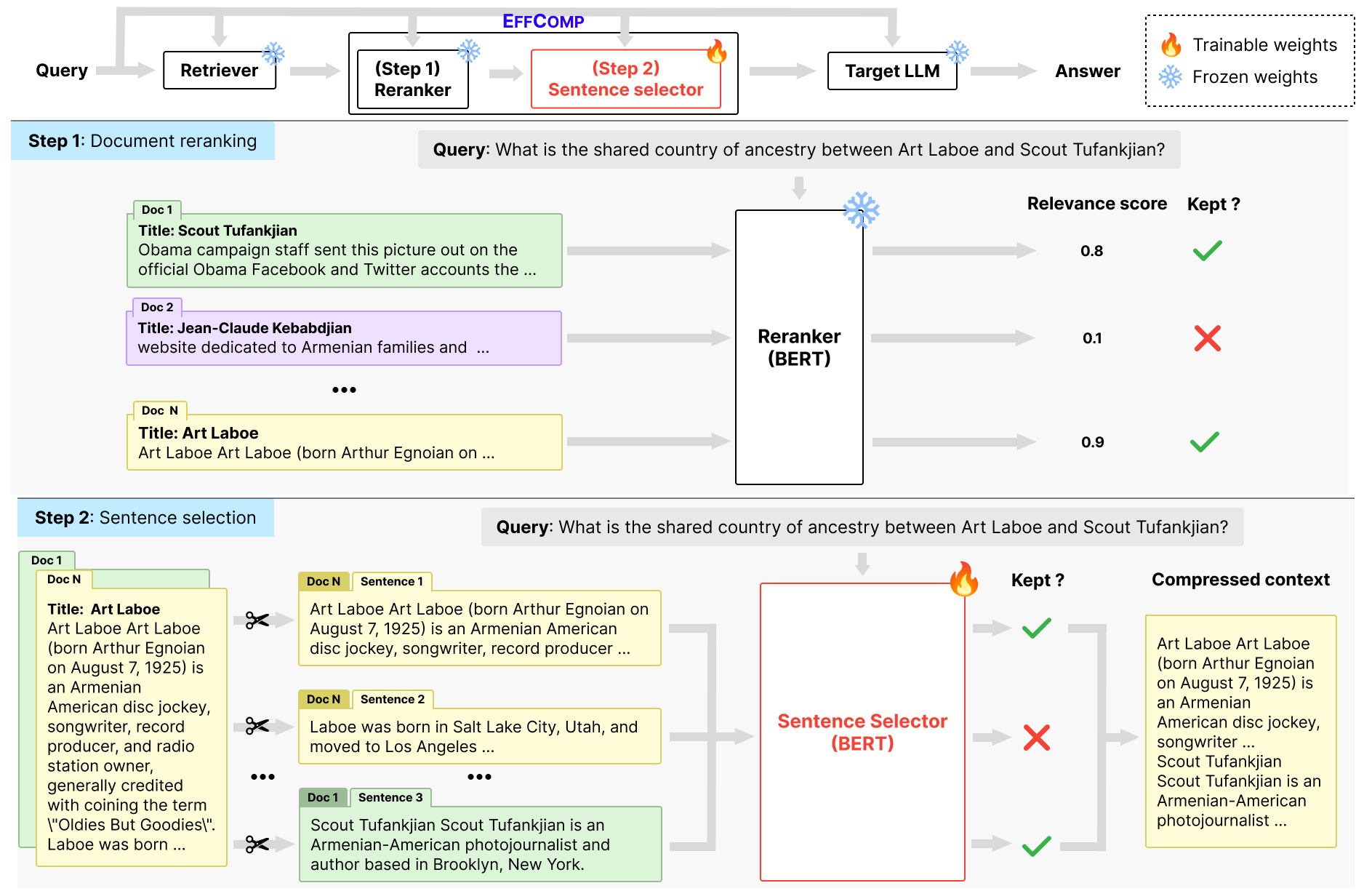}
    \caption{The inference workflow of \met. 
    The top row in the figure illustrates where \met is in the prediction pipeline. 
    In Step~1, retrieved documents are filtered, reordered, and concatenated. In Step~2, they are decomposed into sentences, from which necessary ones are selected and concatenated to form the compressed context for LLM prediction.}
    \label{fig:inference-pipeline}
\end{figure*}

Based on the question $q$, \met compresses the retrieved context in two main steps, one at the document level and the other at the sentence level, as illustrated in Figure~\ref{fig:inference-pipeline}.

\paragraph{Step 1: Document reranking.}
Given $n$ retrieved documents $\langle d_1, d_2, \dots, d_n \rangle$, 
an off-the-shelf document reranker is used to compute a relevancy score for each document $d_i$. The reranker takes the question $q$ and a document $d_i$ as input and outputs a score indicating the relevance of $d_i$ to $q$. Only the top $k$ documents with the highest scores are kept and then sorted by their scores in decreasing order. 
This is performed because evidence appearing earlier in an input prompt is more likely to receive stronger attention from LLMs~\citep{tang-etal-2025-perception}.
We experiment with three BERT-style document rerankers~\citep{devlin2019bertpretrainingdeepbidirectional}, including BGE~\citep{bge_embedding}, Jina~\citep{günther2023jina}, and GTE~\citep{zhang2024mgte}. All of them are relatively lightweight models and can compute the relevancy scores in batches, ensuring high efficiency. 

\paragraph{Step 2: Sentence selection.} \met concatenates the reranked documents from Step 1 and splits the entire concatenated text into a list of sentences $S = \langle s_1, s_2, \ldots, s_m\rangle$. After that, a sentence selector model is applied to identify sentences that should be kept in the final compressed context.
Our sentence selector also relies on the BERT architecture for high efficiency. It takes the question $q$ and all the $m$ sentences in $S$ as input and predicts either 0 or 1 for each sentence $s_i$. Specifically, the input representation $I(q, S)$ at the token level is
``$\texttt{[CLS]}~ q_1 q_2 \ldots q_l ~\texttt{[SEP]} \texttt{[SEN]}~ t_{11} t_{12} \ldots t_{1h_1}$
$~\texttt{[SEN]}~t_{21} \ldots t_{2h_2} ~\ldots \texttt{[SEN]}~ t_{m1} \ldots t_{mh_m} ~\texttt{[SEP]}$'' where $t_{ij}$ is the $j$-th token of $s_i$ 
and $q_{j}$ is the $j$-th token of $q$. 
Apart from the standard \texttt{[CLS]} and \texttt{[SEP]} tokens, we prepend a special \texttt{[SEN]} token to each sentence to support sentence-level classification. 
The sentence selector produces token-level predictions, but only the outputs at the \texttt{[SEN]} tokens are used, indicating whether each corresponding sentence is retained or discarded. The retained sentences are concatenated in order to form the compressed context $\mathbb{C}$. 

Overall, the design of \met offers several advantages.
First, the finest level of selection is at the sentence level (rather than the token level). This renders a compressed context that remains understandable even under extreme compression and aligns naturally with how target LLMs were trained.
Furthermore, the sentence selector in Step 2 considers all remaining context sentences simultaneously, enabling it to discard useful but redundant sentences to maximize the compression ratio.
This simultaneous processing remains efficient thanks to Step 1, which already discards 
a significant number of irrelevant documents.

The remainder of this section explains the training process of the BERT-based sentence selector, which has two phases. 
Phase I first finetunes the model to identify sentences that likely contain the answer. Phase II further finetunes it to select other necessary context sentences and return a concise prompt sufficient for the target LLM.

\subsection{Training Process Phase I: Intermediate Finetuning}
\paragraph{Training set construction.} 
We adapt training examples from a QA dataset to be our supervised finetuning examples.
To explain, each finetuning example consists of a question $q$ from the dataset and a list of retrieved documents $D$ with $|D| = 30$. We concatenate all the documents in $D$, split the concatenated text into sentences $S$, and format them (together with $q$) as input to the sentence selector.
Inspired by document-level label assignment in CompAct~\citep{yoon-etal-2024-compact}, we assign a label to each sentence using the following rule: a sentence is labeled as \textit{retained} if it contains the ground-truth answer string; otherwise, it is labeled \textit{discarded}.
In Phase I, we first train the sentence selector to predict these labels and then focus on additional necessary contexts in Phase II.

\paragraph{Training objective.}  
The training objective is defined at the positions of \texttt{[SEN]} tokens. For each sentence $s_i \in S$, the action (retaining or discarding) is predicted at the position of its preceding \texttt{[SEN]} token. A masked cross-entropy loss is applied such that only \texttt{[SEN]} positions contribute to the objective, while all other tokens are ignored. This encourages the model to learn sentence-level decisions by treating each \texttt{[SEN]} token as the supervision point for its corresponding sentence.

\subsection{Training Process Phase II: Final Finetuning}

The 
model from Phase I focuses only on selecting sentences that contain answers. So, this Phase II aims to further finetune it to select other sentences that may not contain the answers but are necessary for the target LLM to make correct predictions (e.g., bridge sentences in multi-hop reasoning).
We propose using a hybrid reinforcement learning (RL) and supervised learning (SL) approach for Phase II. 
We choose RL, specifically a multi-armed bandit setting, as it is an optimization approach that suits the nature of our sentence selector. 
By design, RL can help further remove redundant and irrelevant context sentences while maintaining the final answer accuracy. 
However, as noted by \citet{shandilya2024tacorltaskawareprompt} and \citet{wei2025reinforcing}, RL fine-tuning under a bandit formulation is computationally expensive and time-consuming because a single non-differentiable reward must be used to optimize multiple actions. 
To overcome this issue, we incorporate SL, which incurs little overhead and generally converges faster because actions are directly supervised.
This results in our hybrid strategy that, for each epoch, updates the model weights using an RL loss followed by an SL loss, as explained next.

\subsubsection{Reinforcement Learning}
We consider our sentence selector as a policy model $\pi_\theta$ and use reward signals derived from the target LLM and ground-truth QA answers to optimize it. 

\paragraph{Problem formulation.} 
We formulate our task as a contextual multi-armed bandit problem~\citep{10535182,pmlr-v9-lu10a}, which terminates after a single decision step. 
A state $\sigma$ is the input $I(q, S)$ to the policy model where $q$ is a question and $S$ is the list of context sentences from the top-$k$ reranked documents. 
The policy $\pi_\theta$ takes $\sigma$ as input and produces output probability distributions for all input tokens. However, consistent with the intermediate finetuning phase, only the 
probability distributions at the \texttt{[SEN]} tokens are used to derive binary actions $a_i \in \{0,1\}$ indicating whether the corresponding sentences are retained (1) or discarded (0). 
With $m$ sentences in $S$, one state produces multiple actions $A = \langle a_1, a_2, \dots, a_m \rangle$ in parallel, where each action is sampled from 
$\text{Categorical }\big(\pi_\theta(\cdot\vert\sigma, i)\big)$
The compressed context $\mathbb{C}$ is then formed by concatenating all retained sentences. 
The target LLM $M$ produces $M(q, \mathbb{C})$, and a reward $r$ for $A$ is then computed.
A trial, therefore, consists of a single state $\sigma$, the full list of binary actions $A$, and the resulting reward $r$.

\paragraph{Reward function.}
We define the reward $r$ as
\[
r =
\begin{cases}
\alpha + (1-\alpha)(1-\tau),
& \text{if Accuracy = 1} \\ 
-(1-\alpha)(1-\tau),
& \text{otherwise}
\end{cases}
\]
where $\tau = |\mathbb{C}|/|\mathbb{D}|$ and the 
Accuracy is obtained by comparing $M(q,\mathbb{C})$ against the ground truth answer of $q$.
This reward jointly maximizes QA accuracy and compression.
During validation, we use it to select the best model checkpoint for evaluation where $\alpha \in (0,1)$ controls the relative importance between the accuracy and the compression ratio. 
In practice, we prefer a relatively high $\alpha$ to avoid extreme compression at the expense of accuracy.

\paragraph{Optimization.}
To train the 
policy $\pi_\theta$,
we adapt 
Group Relative Policy Optimization (GRPO)~\citep{shao2024deepseekmathpushinglimitsmathematical}, a policy gradient method that removes the value function used in PPO~\citep{schulman2017proximalpolicyoptimizationalgorithms} and instead efficiently estimates advantages by normalizing rewards within a group of sampled trials. 
Given a group size $b$, for each state, we sample a group of actions
$\mathbf{A} = \{A_1, \dots, A_b\}$ from $\pi_\theta$, each producing a compressed context $\mathbb{C}_i$
and reward $r_i$. 
The reward set $R = \{r_1, \dots, r_b\}$ is z-normalized in group to yield relative advantages $Z = \{z_1, \ldots, z_b\}$ 
used for updating the policy.

\begin{figure}[t]
    \centering
    \includegraphics[width=1.0\linewidth]{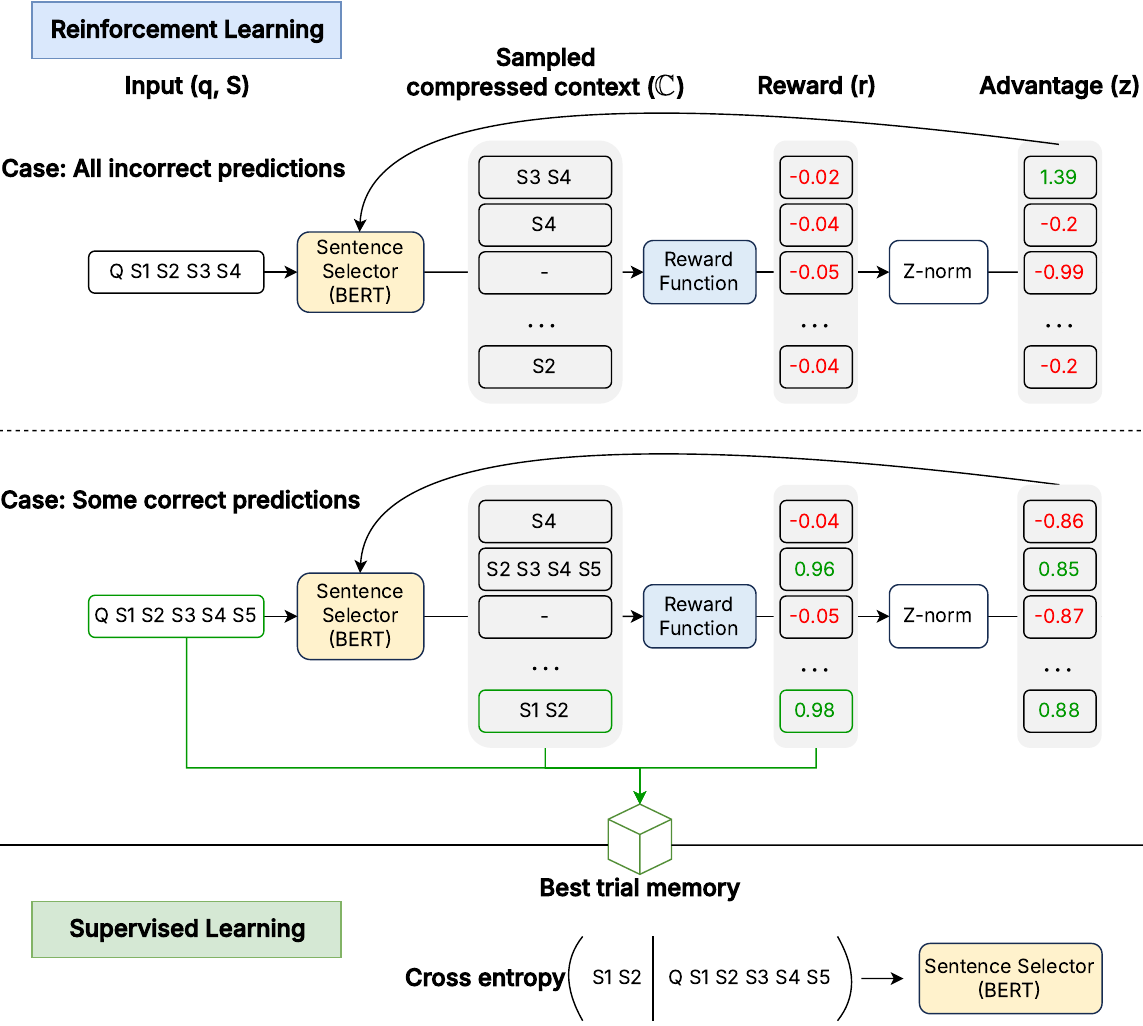}
    \caption{
    A single epoch of the final finetuning (Phase II). In each iteration, we perform reinforcement learning, followed by supervised learning. Reward normalization encourages higher compression when some predictions are correct and lower compression
    otherwise.
    }
    \label{fig:normalization_example}
\end{figure}

Thanks to our reward design, when at least one $A_i$ results in a correct LLM output, the $A_i$ with the highest compression ratio 
receives the greatest advantage, reinforcing the policy to increase compression. Conversely,
if all $A_i \in \mathbf{A}$ lead to incorrect LLM outputs, the $A_i$ with the lowest compression ratio is assigned the highest advantage, incentivizing the policy to reduce compression and preserve more information. See Figure~\ref{fig:normalization_example} for an illustration.

\paragraph{Training set construction.}
For each QA training example $(q, y_q)$, we perform Step 1 (document reranking) to obtain context sentences $S$ from the top-$k$ documents to form the input $I(q, S)$. 
However, we exclude some training examples.
First, we exclude examples that the target LLM cannot answer correctly even with all sentences in $S$ provided, as further compression is unlikely to help.
Given our reward function, including such examples would bias the policy to always retain all sentences, making the training inefficient.
Second, we exclude examples that the target LLM can answer correctly even without any context. Keeping these examples would 
encourage the policy to discard all the sentences,
resulting in non-grounded LLM outputs.
This example filtering step ensures that
the training set only
contains examples where context is necessary and compression choices truly matter. 

\subsubsection{Supervised Learning}
During GRPO, we sample multiple $A_i$ for each state $\sigma$. To enhance training efficiency, we introduce a memory $C$ to store the best trial of each state observed so far (i.e., the best list of sentence-level binary actions so far for each example).
For each state $\sigma$, we begin storing a trial $(\sigma, A_i, r_i)$ in $C$ when $r_i > 0$ and update it when we find another trial $(\sigma, A_j, r_j)$ where $r_j \geq r_i$.
After the RL update in each epoch, these best trials in $C$ are used as silver labels to finetune the policy (i.e., the sentence selector) in a supervised learning manner.
This concludes the epoch.
Then we begin the next iteration with RL and SL again.
We refer readers to Appendix \ref{sec:algorithm} for the detailed algorithm.

\begin{table*}[t]
\centering
\footnotesize
\setlength{\tabcolsep}{3pt}
\renewcommand{\arraystretch}{1.25}

\begin{tabular}{l|rccc|rccc|rccc|rccc}
\hline
\textbf{Method}
& \multicolumn{4}{c|}{\textbf{HQA}}
& \multicolumn{4}{c|}{\textbf{2WikiQA}}
& \multicolumn{4}{c|}{\textbf{NQ}}
& \multicolumn{4}{c}{\textbf{TQA}} \\
\cline{2-17}
& $1/\tau$ & Acc & $\mathrm{C}_{0.5}$ & $\mathrm{C}_{1}$
& $1/\tau$ & Acc & $\mathrm{C}_{0.5}$ & $\mathrm{C}_{1}$
& $1/\tau$ & Acc & $\mathrm{C}_{0.5}$ & $\mathrm{C}_{1}$
& $1/\tau$ & Acc & $\mathrm{C}_{0.5}$ & $\mathrm{C}_{1}$\\
\hline
Raw Documents
& 1.0x & 34.6  & 0.00 & 0.00
& 1.0x & 29.1  & 0.00 & 0.00
& 1.0x & 30.1  & 0.00 & 0.00
& 1.0x & 77.4  & 0.00 & 0.00\\
\hline
\multicolumn{17}{l}{\textit{Reranker-only methods}} \\
BGE reranker & 3.0x & 35.2  & 0.39 & 0.46  & 3.0x & 23.8  & 0.27 & 0.35 & 3.0x & 29.6  & 0.33 & 0.41 & 3.0x & 79.9  & 0.77 & 0.73\\
Jina reranker & 3.0x & 34.6  & 0.38 & 0.46 & 3.0x & 24.4  & 0.28 & 0.36 & 3.0x & 30.6  & 0.34 & 0.42 & 3.0x & 80.1   & 0.77 & 0.73\\
GTE reranker & 3.0x & 34.7  & 0.38 & 0.46 & 3.0x & 23.9  & 0.27 & 0.35 & 3.0x & 30.6  & 0.34 & 0.42 & 3.0x & 80.4   & 0.77 & 0.73\\
\hline
\multicolumn{17}{l}{\textit{Compression frameworks}} \\
LLMLingua2 & 5.4x & 25.0   & 0.29 & 0.38 & 5.5x & 19.8  & 0.23 & 0.32 & 5.3x & 23.8 & 0.28 & 0.37 & 5.4x & 72.9 & 0.75 & 0.77 \\
DAC & 5.1x & 24.7  & 0.29 & 0.38 & 5.0x & 21.0  & 0.25 & 0.33 & 5.2x & 22.7  & 0.27 & 0.35 & 5.2x & 73.5 & 0.75 & 0.77 \\
CPC & 5.6x & 30.5 & 0.35 & 0.45 & 5.9x & 21.8 & 0.26 & 0.35 & 5.2x & 28.0 & 0.32 & 0.42 & 5.4x & 78.6 & 0.79 & 0.80 \\
LongLLMLingua & 5.1x & 31.8 & 0.36 & 0.46 & 5.1x & 21.8 & 0.26 & 0.34 & 5.1x & 25.7 & 0.30 & 0.39 & 5.1x & 78.4 & 0.79 & 0.79 \\
Recomp$_{\mathrm{Extractive}}$ & 29.2x & 25.6 & 0.30 & 0.40 & 29.3x & 17.4 & 0.21 & 0.30 & 29.8x & 22.8 & 0.27 & 0.37 & 32.0x & 73.1 & 0.77 & 0.83 \\
Recomp$_{\mathrm{Abstractive}}$ & \textbf{132.1x} & 27.9 & 0.33 & 0.44 & \textbf{126.9x} & 21.0 & 0.25 & 0.35 & \textbf{48.3x} & 27.6 & 0.32 & 0.43 & \textbf{139x} & 73.0 & 0.77 & 0.84 \\
CompAct & 48.7x & 31.2 & 0.36 & 0.47 & 54.7x & 16.8 & 0.20 & 0.29 & 44.2x & 27.8 & 0.32 & 0.43 & 50.2x & 77.6 & 0.81 & 0.87 \\
\hline
\multicolumn{17}{l}{\textit{Our work: \met}} \\
No reranker & 26.8x & \textbf{37.4} & \textbf{0.43} & \textbf{0.54} & 11.7x & \textbf{28.9} & \textbf{0.34} & \textbf{0.44}
& 9.8x & \textbf{31.1} & \textbf{0.36} & \textbf{0.46}
& 32.8x & 78.9 & 0.82 & 0.87
\\
\hdashline
BGE reranker  & 39.5x & 36.3 & 0.42 & 0.53 & 19.4x & 25.0 & 0.29 & 0.40
& 14.1x & 29.0 & 0.34 & 0.44
& 49.6x & 79.5 & \textbf{0.83} & \textbf{0.88}\\
Jina reranker & 38.5x & 35.8 & 0.41 & 0.52 & 19.3x & 25.8 & 0.30 & 0.41
& 18.8x & 29.7 & 0.34 & 0.45
& 50.0x & 79.6 & \textbf{0.83} & \textbf{0.88}\\
GTE reranker & 40.4x & 35.5 & 0.41 & 0.52 & 20.4x & 25.0 & 0.29 & 0.40
& 19.2x & 29.7 & 0.34 & 0.45
& 54.1x & \textbf{79.8} & \textbf{0.83} & \textbf{0.88}\\
\hline
\end{tabular}

\caption{
Results on four QA benchmarks for Gemma-2-9B-IT.
We report compression ratio ($1/\tau$), accuracy, and $\mathrm{C}_{\beta}$ for $\beta \in \{0.5,1\}$. Higher is better.  
Best values are in bold. 
}
\label{tab:qa-result-gemma}
\end{table*}

\section{Experiments}
\label{sec:experiments}

We trained \met on the HotpotQA training set and evaluated it on four open-domain QA datasets: HotpotQA~(HQA; \citet{yang2018hotpotqa}), 2WikiMultiHopQA~(2WikiQA; \citet{xanh2020_2wikimultihop}), Natural Questions~(NQ; \citet{47761}), and TriviaQA~(TQA; \citet{2017arXivtriviaqa}).

\subsection{Experimental Setup}
\paragraph{Implementation details.}
Following prior work~\citep{yoon-etal-2024-compact,xu2024recomp}, we implemented the document retrieval system using Contriever~\citep{izacard2021contriever}, finetuned on MSMARCO~\citep{bajaj2018msmarcohumangenerated}. 
For each question, we retrieved $n=30$ documents from the 2018 Wikipedia corpus~\citep{karpukhin-etal-2020-dense}. Additionally, we used Gemma-2-9B-IT\footnote{https://huggingface.co/google/Gemma-2-9B-IT}~\citep{gemma_2024} as the target LLM. 
For \met, we tried three document rerankers, including BGE\footnote{https://huggingface.co/BAAI/bge-reranker-base}~\citep{bge_embedding}, Jina\footnote{https://huggingface.co/jinaai/jina-reranker-v2-base-multilingual}~\citep{günther2023jina}, and GTE\footnote{https://huggingface.co/Alibaba-NLP/gte-reranker-modernbert-base}~\citep{zhang2024mgte}, to rerank and select the top-$k$ documents ($k=10$) in Step 1 of Figure~\ref{fig:inference-pipeline}.
For Step 2, we used ModernBERT-base\footnote{https://huggingface.co/answerdotai/ModernBERT-base}~\citep{modernbert} as the sentence selector and used the BGE reranker to select documents as input for finetuning the sentence selector. 
We also evaluated \met without Step 1, which passed sentences from $n$ documents to Step 2 without reranking. This variant is denoted as ``No reranker'' in the result tables. 

\paragraph{Baselines.}
We compared \met against several types of baselines.
The simplest one was ``Raw Documents'', which inputs the $n$ retrieved documents to the target LLM without compression.
Next, reranker-only filtering baselines used a reranker (BGE, Jina, or GTE) to select top-$k$ documents to feed into the target LLM. This is equivalent to \met without Step 2.
Furthermore, we compared \met with state-of-the-art compression frameworks, including five extractive methods (Recomp-Extractive~\citep{xu2024recomp}, LLMLingua2~\citep{pan-etal-2024-llmlingua}, DAC~\citep{zhao-etal-2025-dac}, CPC~\citep{Liskavets_Ushakov_Roy_Klibanov_Etemad_Luke_2025}, LongLLMLingua~\citep{jiang-etal-2024-longllmlingua}) and two abstractive methods (Recomp-Abstractive~\citep{xu2024recomp} and CompAct~\citep{yoon-etal-2024-compact}).
Implementation
details of the baselines are provided in Appendix~\ref{sec:baseline-detail}.

\paragraph{Metrics.}
We evaluate all methods using three metrics: compression ratio ($1/\tau$), accuracy (Acc), and $\mathrm{C}_{\beta}$ score. The $\mathrm{C}_{\beta}$ score combines compression ratio and accuracy into a single metric for easier comparison. It is defined as the weighted harmonic mean between accuracy and compression rate:\footnote{The compression rate $(1-1/\tau)$ is a normalized transformation of the compression ratio into the range $[0,1]$, enabling it to be combined with accuracy in the harmonic mean.}
\[
\mathrm{C}_{\beta}
=
\frac{(1+\beta^2)\cdot \mathrm{Acc}\cdot (1-\tau)}
{
\beta^2 \cdot \mathrm{Acc} + (1-\tau)
}
\]
where $\beta$ is the weighting factor. Lower $\beta$ places more emphasis on accuracy. We report $\mathrm{C}_{0.5}$, which places greater emphasis
on accuracy, and $\mathrm{C}_{1}$, which equally balances accuracy and
compression.

\subsection{Results}
Table~\ref{tab:qa-result-gemma} reports the results on four datasets. \met consistently achieved the highest $\mathrm{C}_{\beta}$ scores across all baselines, demonstrating a superior balance between compression efficiency and QA performance. Notably, although \met was trained only on HQA, it maintained strong performance on the out-of-distribution datasets, including 2WikiQA, NQ, and TQA. This indicates that the learned sentence selection strategy generalizes effectively beyond the training dataset rather than being specific to HQA. The improvements are especially notable on multi-hop QA datasets such as HQA and 2WikiQA, where preserving supporting evidence is crucial. For example, on HQA, \met achieved a $\mathrm{C}_{1}$ score of 0.54, representing an approximately 15\% relative improvement over the strongest prior compression framework. Compared to using the Raw Documents, \met with rerankers achieved competitive or sometimes higher Acc with compression ratios of 14.1x--54.1x. For example, on HQA, \met with BGE reranker achieved 36.3\% accuracy with the compression ratio of 39.5x, while the Raw Documents got 34.6\% accuracy with no compression (1x). This demonstrates that context compression, if done appropriately, can enhance rather than degrade the quality of LLM predictions. Without the rerankers, \met got significantly lower compression ratios but usually slightly higher Acc, resulting in higher $\mathrm{C}_{\beta}$. 
Compared to reranker-only methods, \met with the same reranker achieved significantly higher compression ratios (e.g., from 3x to 32.8x--54.1x on TQA) and competitive or higher Acc scores.
This highlights the immense impact of the sentence selection step of \met on compression ratios and indicates that our sentence selector, trained on BGE reranked data, generalized effectively to other rerankers such as Jina and GTE.
Furthermore, \met outperformed all other compression frameworks in terms of accuracy and $\mathrm{C}_{\beta}$, suggesting a better balance between the compression ratios and the QA quality.

\section{Additional Analyses}
In this section, we conduct additional analyses to better understand \met. Specifically, we evaluate \met on additional QA datasets and with commercial LLMs. We also analyze the contributions of the two finetuning phases to sentence selection and compare our hybrid RL-SL training strategy with alternative training strategies.

\label{sec:analysis}

\begin{table*}[t]
\centering
\footnotesize
\setlength{\tabcolsep}{4pt}
\renewcommand{\arraystretch}{1.2}

\begin{tabular}{l|rcccc|rcccc|rcccc}
\hline
\multicolumn{16}{c}{\textbf{Multi-hop}} \\
\hline
\textbf{Method}
& \multicolumn{5}{c|}{\textbf{HQA}}
& \multicolumn{5}{c|}{\textbf{2WikiQA}}
& \multicolumn{5}{c}{\textbf{MuSiQue}} \\
\cline{2-16}
& $1/\tau$ & Acc & F1 & $\mathrm{C}_{0.5}$ & $\mathrm{C}_{1}$
& $1/\tau$ & Acc & F1 & $\mathrm{C}_{0.5}$ & $\mathrm{C}_{1}$
& $1/\tau$ & Acc & F1 & $\mathrm{C}_{0.5}$ & $\mathrm{C}_{1}$\\
\hline
Raw Documents   & 1.0x & 36.5 & 46.6 & 0.00 & 0.00 & 1.0x & 40.5 & 43.7 & 0.00 & 0.00 & 1.0x & 20.5 & 26.5 & 0.00 & 0.00 \\
DAC             & 6.5x & 25.0 & 34.9 & 0.29 & 0.39 & 5.4x & 24.0 & 28.4 & 0.28 & 0.37 & 6.7x & 7.0 & 11.8 & 0.09 & 0.13 \\
LLMLingua2      & 5.4x & 33.0 & 42.0 & 0.37 & 0.47 & 5.7x & 27.0 & 30.3 & 0.31 & 0.41 & 5.5x & 16.0 & 21.8 & 0.19 & 0.27 \\
LongLLMLingua   & 3.5x & 32.0 & 41.9 & 0.36 & 0.44 & 3.5x & 38.5 & 43.9 & 0.42 & 0.5 & 3.5x & 15.5 & 21.1 & 0.18 & 0.25 \\
CPC             & 5.1x & 35.0 & 45.6 & 0.39 & 0.49 & 5.2x & 38.5 & 43.2 & 0.43 & 0.52 & 5.1x & 17.0 & 21.5 & 0.20 & 0.28 \\
\hline
\multicolumn{16}{l}{\textit{Our work:} \met} \\
Ours (5x) & 5.2x & 40.0 & 50.1 & 0.44 & 0.54 & 5.1x & 43.0 & 47.0 & 0.47 & 0.56 & 5.1x & 22.0 & 29.3 & 0.26 & 0.35 \\
Ours (No reranker) & 39.3x & 35.5 & 46.1 & 0.41 & 0.52 & 13.0x & 43.0 & 47.3 & 0.48 & 0.59 & 13.3x & \textbf{23.5} & \textbf{29.8} & \textbf{0.28} & \textbf{0.37} \\
\hdashline
Ours (BGE) & \textbf{65.8x} & \textbf{40.5} & 49.0 & \textbf{0.46} & \textbf{0.57} & 22.5x & \textbf{50.0} & \textbf{53.7} & \textbf{0.55} & \textbf{0.66} & 40.3x & 20.0 & 26.2 & 0.24 & 0.33 \\
Ours (GTE) & 58.5x & \textbf{40.5} & \textbf{50.5} & \textbf{0.46} & \textbf{0.57} & 21.9x & 48.0 & 52.2 & 0.53 & 0.64 & \textbf{43.2x} & 18.0 & 25.5 & 0.22 & 0.30 \\
Ours (Jina) & 63.4x & 38.0 & 47.7 & 0.43 & 0.55 & \textbf{25.7x} & 47.5 & 51.5 & 0.53 & 0.64 & 32.6x & 22.5 & 28.1 & 0.27 & \textbf{0.37} \\
\hline

\multicolumn{16}{c}{\textbf{Single-hop}} \\
\hline
\textbf{Method}
& \multicolumn{5}{c|}{\textbf{MultiFieldQA}}
& \multicolumn{5}{c|}{\textbf{NarrativeQA}}
& \multicolumn{5}{c}{\textbf{Qasper}} \\
\cline{2-16}
& $1/\tau$ & Acc & F1 & $\mathrm{C}_{0.5}$ & $\mathrm{C}_{1}$
& $1/\tau$ & Acc & F1 & $\mathrm{C}_{0.5}$ & $\mathrm{C}_{1}$
& $1/\tau$ & Acc & F1 & $\mathrm{C}_{0.5}$ & $\mathrm{C}_{1}$\\
\hline
Raw Documents   & 1.0x & 14.0 & 40.6 & 0.00 & 0.00 & 1.0x & \textbf{12.5} & \textbf{23.6} & 0.00 & 0.00 & 1.0x & 22.0 & 39.6 & 0.00 & 0.00 \\
DAC             & 5.9x & 4.7 & 26.9 & 0.06 & 0.09 & 7.2x & 4.5 & 13.7 & 0.06 & 0.09 & 5.0x & 15.5 & 24.3 & 0.18 & 0.26 \\
LLMLingua2      & 5.6x & 10.7 & 31.1 & 0.13 & 0.19 & 6.0x & 12.0 & 20.8 & 0.14 & \textbf{0.21} & \textbf{5.2x} & 18.5 & 28.7 & 0.22 & 0.30 \\
LongLLMLingua   & 4.4x & 12.0 & 33.7 & 0.14 & 0.21 & 3.6x & 10.5 & 19.6 & 0.13 & 0.18 & 3.3x & 16.0 & 22.3 & 0.19 & 0.26 \\
CPC             & 5.1x & 20.0 & 45.3 & 0.24 & 0.32 & 5.5x & 8.5 & 20.0 & 0.1 & 0.15 & 5.1x & 19.5 & 34.9 & 0.23 & 0.31 \\
\hline
\multicolumn{16}{l}{\textit{Our work:} \met} \\
Ours (5x) & 5.0x & \textbf{21.3} & \textbf{48.3} & \textbf{0.25} & \textbf{0.34} & 5.4x & 10.5 & 18.6 & 0.13 & 0.19 & 5.1x & 22.5 & 39.0 & 0.26 & \textbf{0.35} \\
Ours (No reranker) & 24.3x & 19.3 & 45.8 & 0.23 & 0.32 & 1.6x & 12.0 & 22.1 & 0.14 & 0.18 & 3.0x & \textbf{24.0} & \textbf{40.2} & \textbf{0.28} & \textbf{0.35} \\
\hdashline
Ours (BGE) & 28.2x & 20.0 & 44.4 & 0.24 & 0.33 & 22.5x & 10.5 & 18.2 & 0.13 & 0.19 & 3.0x & \textbf{24.0} & \textbf{40.2} & \textbf{0.28} & \textbf{0.35} \\
Ours (GTE) & \textbf{32.3x} & 19.3 & 45.7 & 0.23 & 0.32 & \textbf{26.1x} & 9.5 & 16.1 & 0.12 & 0.17 & 3.0x & \textbf{24.0} & \textbf{40.2} & \textbf{0.28} & \textbf{0.35} \\
Ours (Jina) & 28.6x & 18.7 & 44.2 & 0.22 & 0.31 & 20.3x & 12.0 & 19.6 & \textbf{0.15} & \textbf{0.21} & 3.0x & \textbf{24.0} & \textbf{40.2} & \textbf{0.28} & \textbf{0.35} \\
\hline

\end{tabular}

\caption{
Results on LongBench QA datasets (out of distribution) using Gemma-2-9B-IT as the target LLM. The baselines are evaluated at a 5x compression ratio. We report compression ratio ($1/\tau$), accuracy, token level F1 score, and $\mathrm{C}_{\beta}$. Higher is better, and best values are in bold.
}
\label{tab:longbench-gemma}
\end{table*}

\paragraph{RQ1: How does \met perform on the Longbench benchmark?}
We investigate how \met performs in real-world out-of-distribution scenarios using the HQA-trained version of our compressor model. To do so, we use the QA (English) subset of the LongBench benchmark~\citep{bai-etal-2024-longbench},
which includes both multi-hop and single-hop QA datasets. The multi-hop
datasets consist of HQA, 2WikiQA, and MuSiQue,
which require reasoning over multiple pieces of evidence. In contrast, the
single-hop datasets represent a more distinct out-of-distribution setting
from the HQA training data and cover diverse long-context domains:
MultiFieldQA requires answering questions based on long articles,
NarrativeQA focuses on questions about stories and scripts, and Qasper
requires answering questions based on NLP research papers. We truncate the
context length to match the maximum input length of Gemma-2-9B-IT. Although \met is designed as an adaptive compression method that dynamically determines the compression level for each example, for this research
question we additionally evaluate it under fixed compression settings for fair comparison with prior extractive baselines. For fixed compression, we use an \met variant that selects the top-ranked sentences to achieve the target compression ratio. We report the compression ratio, Acc, F1, and $\mathrm{C}_{\beta}$. For reranker-based variants, we first use the reranker to select the top 10 documents before applying sentence-level compression. In Table~\ref{tab:longbench-gemma}, \met achieves the best $\mathrm{C}_{\beta}$ scores across all multi-hop settings compared to the other baselines and outperforms all compression baselines in single-hop settings. Furthermore, reranker-based \met achieves the highest $\mathrm{C}_{\beta}$ scores in five out of six evaluated datasets, highlighting the importance of document-level filtering.

\begin{figure*}[h]
    \small
    \centering
    \begin{minipage}{0.44\linewidth}
        \includegraphics[width=\linewidth]{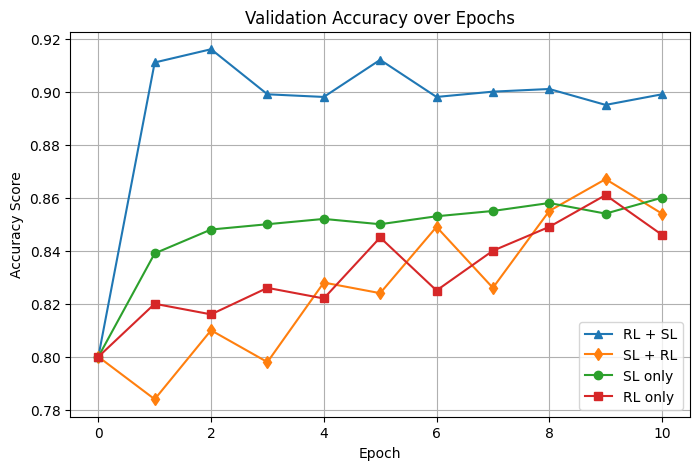}
    \end{minipage}
    \begin{minipage}{0.44\linewidth}
        \includegraphics[width=\linewidth]{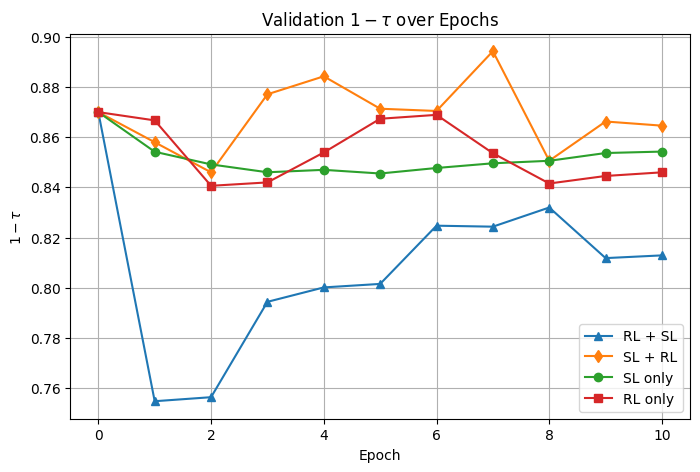}
    \end{minipage}
    
    \caption{Ablation study of different training strategies for \met\ on HQA. 
    The plots compare (left) validation accuracy and (right) compression rate $1-\tau = 1 - |\mathbb{C}|/|\mathbb{D}|$ during validation across ten epochs. For both metrics, the higher values are better.}
    \label{fig:more-training-ablation}
\end{figure*}

\paragraph{RQ2: How does \met perform with commercial LLMs?}
To assess whether \met generalizes to larger commercial LLMs, we further evaluate \met with larger commercial LLMs, including GPT-4o mini\footnote{https://developers.openai.com/api/docs/models/gpt-4o-mini} and Claude Sonnet 4.5\footnote{https://platform.claude.com/docs/en/models/sonnet-4-5/overview}, using temperature 0. We apply the same \met model reported in Table~1 to the first 500 examples of the HQA test set. The results are presented in Table~\ref{tab:commercial-llm}.

\begin{table}[t]
\centering
\footnotesize
\setlength{\tabcolsep}{5pt}
\renewcommand{\arraystretch}{1.2}
\begin{tabular}{lrccc}
\hline
\textbf{Method}
& $1/\tau$ & Acc & $\mathrm{C}_{0.5}$ & $\mathrm{C}_{1}$ \\
\hline

\multicolumn{5}{l}{\textbf{Target LLM: Claude Sonnet 4.5}} \\
Raw Documents
& 1.0x & 28.0 & 0.00 & 0.00 \\
LLMLingua2
& 5.4x & 40.2 & 0.45 & 0.54 \\
DAC
& 5.1x & 40.2 & 0.45 & 0.54 \\
CPC
& 5.6x & 42.4 & 0.47 & 0.56 \\
LongLLMLingua
& 5.1x & 43.6 & 0.48 & 0.57 \\
Recomp Extractive
& 28.5x & 36.8 & 0.42 & 0.53 \\
Recomp Abstractive
& 134.7x & 31.4 & 0.36 & 0.48 \\
CompAct
& 49.0x & 36.4 & 0.42 & 0.53 \\
\hline
\multicolumn{5}{l}{\textit{Our work:} \met} \\
\met (No reranker)
& 28.0x & 41.6 & 0.47 & 0.58 \\
\hdashline
\met (BGE reranker)
& \textbf{43.6x} & \textbf{45.0} & \textbf{0.50} & \textbf{0.62} \\

\hline
\multicolumn{5}{l}{\textbf{Target LLM: GPT-4o mini}} \\
Raw Documents
& 1.0x & 35.8 & 0.00 & 0.00 \\
LLMLingua2
& 5.4x & 30.4 & 0.35 & 0.44 \\
DAC
& 5.1x & 27.0 & 0.31 & 0.40 \\
CPC
& 5.6x & 32.8 & 0.37 & 0.47 \\
LongLLMLingua
& 5.1x & 35.8 & 0.40 & 0.50 \\
Recomp Extractive
& 28.5x & 28.6 & 0.33 & 0.44 \\
Recomp Abstractive
& 134.7x & 28.6 & 0.33 & 0.44 \\
CompAct
& 49.0x & 32.6 & 0.38 & 0.49 \\
\hline
\multicolumn{5}{l}{\textit{Our work:} \met} \\
\met (No reranker)
& 28.0x & \textbf{36.2} & \textbf{0.41} & \textbf{0.53} \\
\hdashline
\met (BGE reranker)
& 43.6x & 35.0 & 0.40 & 0.52 \\

\hline
\end{tabular}
\caption{Performance of prompt compression methods with commercial LLMs on HQA.}
\label{tab:commercial-llm}
\end{table}

These results demonstrate that \met generalizes effectively across commercial LLMs in terms of $\mathrm{C}_{\beta}$. With Claude Sonnet 4.5, \met with the BGE reranker achieves the highest $\mathrm{C}_{0.5}$ and $\mathrm{C}_{1}$ scores of 0.50 and 0.62, respectively, while attaining a 43.6x compression ratio. With GPT-4o mini, \met without reranking achieves the highest $\mathrm{C}_{0.5}$ and $\mathrm{C}_{1}$ scores of 0.41 and 0.53, respectively, while the BGE-reranked variant remains competitive with $\mathrm{C}_{0.5}=0.40$ and $\mathrm{C}_{1}=0.52$ at a higher compression ratio of 43.6x. Overall, these results show that \met maintains a strong trade-off between compression and QA performance across different commercial target LLMs. This generalization possibly stems from our training set construction (Section~\ref{sec:method}), which uses highly informative examples identified by a relatively small LLM that generalize effectively to different target LLMs.


\paragraph{RQ3: How do the two finetuning phases contribute to the accuracy of sentence selection?}
This RQ investigates the effects of Phase I (intermediate finetuning) and Phase II (final finetuning) on the quality of compressed contexts. Specifically, we conduct an ablation study by training our sentence selector under three configurations: Phase I only, Phase II only, and our proposed Phase I + Phase II.
We evaluate these models on HQA, a multi-hop QA dataset that provides gold labels specifying supporting context sentences required to answer a question. We measure the recall of these supporting sentences by checking whether at least 50\% of the tokens from each supporting sentence appear as a subsequence in the compressed output.
The results show that the Phase I-only model achieves a supporting fact recall of 75.83\%, while the Phase II-only model achieves 84.13\%. Combining both phases further increases the supporting fact recall to 91.09\%. 
These results confirm that while Phase I focuses primarily on sentences containing the direct answers, Phase II significantly helps the model retrieve critical "bridge" sentences, thereby improving its effectiveness in multi-hop reasoning.

\paragraph{RQ4: How does the hybrid RL-SL finetuning approach perform compared to other approaches?}
We conduct an ablation study on the HQA dataset to evaluate four finetuning strategies: (1) reinforcement learning (RL) only, (2) supervised learning (SL) only using the best sampled trials, (3) the hybrid approach that performs SL then RL in each iteration (SL+RL) and (4) the hybrid approach of \met that performs RL then SL in each iteration (RL+SL). 
Figure~\ref{fig:more-training-ablation} shows that, overall, the RL only and SL + RL strategies primarily optimize compression, maintaining competitive accuracy even at high compression rates, but require more training epochs to converge. 
Meanwhile, the SL only strategy generally attains higher validation accuracy than the RL only approach, but exhibits a lower and more stable compression rate across epochs, suggesting a more limited search space.
Our hybrid (RL+SL) strategy, in contrast, achieves notably higher validation accuracy during the early epochs, while exhibiting a relatively lower compression rate. Still, the compression rate is notably higher than several extractive compression methods as seen in Table~\ref{tab:qa-result-gemma}.
In conclusion, our hybrid (RL+SL) approach leads to a more accurate LLM output sooner than the non-hybrid approaches, demonstrating the efficiency of our training framework.

\section{Related Work}

\paragraph{Prompt compression.} 
\citet{li-etal-2025-prompt} categorize prompt compression methods into two types.
Soft prompt methods compress inputs into continuous vectors readable only by the target LLM~\citep{wingate-etal-2022-prompt,chevalier-etal-2023-adapting,mu2023learning,ge2024incontext}.
In contrast, hard prompt methods use truncation, selection, or summarization to keep the compressed prompts human-readable \citep{li-etal-2023-compressing,jiang-etal-2024-longllmlingua,xu2024recomp,yoon-etal-2024-compact,fei2025efficient}. \met falls into the latter category as we value interpretability and applicability across different target LLMs.
So far, few hard prompt methods have employed reinforcement learning. 
\citet{10535182} formulate prompt compression as a token-level contextual bandit problem for instruction-following tasks. \citet{shandilya2024tacorltaskawareprompt} propose TACO-RL for QA and summarization. 
They employ a constant penalty to counter the overly aggressive compression often encouraged by its optimization objectives.
\met, by contrast, operates at the sentence level with a tailored reward design and training set construction.
This naturally regulates compression without the fixed limits, allowing higher compression while preserving QA accuracy.

\paragraph{Combining reinforcement learning (RL) and supervised finetuning (SFT).}
Recent studies show that using SFT to train LLMs followed by RL to align them with desired behaviors effectively enhances their capabilities~\citep{deepseekai2025deepseekr1incentivizingreasoningcapability,ouyang2022traininglanguagemodelsfollow}. Instead of using one after the other, SuperRL~\citep{liu2025superrlreinforcementlearningsupervision} enhances the complex reasoning ability of LLMs using RL. However, when all sampled trials of an example yield zero reward, leading to uninformative RL signals, they switch to SFT with high-quality offline data.
In contrast, our framework focuses on improving the efficiency of the training process while alleviating the zero-reward issue by using a dense reward and setting an RL hyperparameter (Section~\ref{sec:training-configuration}) to encourage the exploration of various action choices.
It then uses the best-performing trials from RL (instead of offline data) for SFT before alternating between RL and SFT in subsequent epochs.

\section{Conclusion}
We propose \met, a prompt compression framework for RAG-based QA.
\met performs context selection at the document level and then the sentence level, rendering minimal yet sufficient context for the target LLM.
Our hybrid finetuning achieves the best overall training performance in terms of accuracy and consistently obtains the strongest $\mathrm{C}_{\beta}$ scores among all baselines.
Experimental results show that \met yields high compression ratios (up to 54.1x) while maintaining or even improving the QA accuracy (up to 8.1\% relative to the no-compression baseline). 

\section*{Limitations}
We acknowledge that \met has some limitations.
First, \met was designed to tackle RAG-based question answering, so its algorithm may not work effectively on other tasks such as text summarization.
Second, \met is constrained by the context length of the underlying BERT-based encoder, which is currently limited to 8,192 tokens in state-of-the-art implementations. While \met uses a reranker to fit the input into a fixed-size window, the method is still fundamentally bounded by the encoder’s context-length limitation.
Third, while adaptive compression improves QA accuracy by dynamically preserving more informative content, it also produces variable compressed context lengths, which may lead to unstable or unexpectedly high inference costs in practical deployments. This limitation could potentially be mitigated by incorporating fixed-rate compression as we did in RQ1.



\bibliography{custom}

\onecolumn
\appendix

\section{Detailed Algorithm}
\label{sec:algorithm}

\subsection{Reinforcement learning}

This section provides an in-depth description of the RL finetuning procedure. 
To train the policy $\pi_\theta$, 
we modify GRPO by excluding the explicit KL penalty from the loss function.
Instead, we monitor the approximate KL divergence~\citep{schulman2020kl} between the updated policy and the reference policy during optimization and terminate updates early if it exceeds a threshold.

For each state $\sigma$, the policy produces multiple actions in parallel, denoted by
$A = \langle a_1, a_2, \dots, a_m \rangle$, where each action is independently sampled from a categorical distribution:
\begin{equation}
a_i \sim \text{Categorical }\!\big(\pi_\theta(\cdot \mid \sigma, i)\big),
\quad \forall i \in \{1,\dots,m\}.
\label{eq:sampling-categorical}
\end{equation}

To measure the degree of policy change,
we define 
$\pi_\theta(A \mid \sigma)$
as the geometric mean of the probabilities across all 
sampled binary
actions in $A$:
\begin{equation}
\pi_\theta(A \mid \sigma) = \left( \prod_{i=1}^m \pi_\theta(a_{i} \mid \sigma) \right)^{\tfrac{1}{m}}
\label{eq:grpo-geommean}
\end{equation}

Our reinforcement learning objective is then defined as
\begin{equation}
\mathcal{L}_{\text{RL}}(\theta) = 
-\mathbb{E}
\left[
\min\!\left( 
\frac{\pi_\theta(A \mid \sigma)}{\pi_{\theta_{\text{old}}}(A \mid \sigma)} z,\;
\text{clip}\!\left(
\frac{\pi_\theta(A \mid \sigma)}{\pi_{\theta_{\text{old}}}(A \mid \sigma)},\; 1-\epsilon,\; 1+\epsilon
\right) z
\right)
\right],
\label{eq:rl-clip}
\end{equation}
where $\pi_\theta$ and $\pi_{\theta_{\text{old}}}$ denote the current and old policy models, respectively, and $\epsilon$ is a clipping hyperparameter, typically set to $0.2$.

Additionally, to quantify the degree of action exploration, we compute the Shannon entropy~\citep{shannon1948mathematical} of the policy $\pi_\theta$ at the state $\sigma$ (considered only at the \texttt{[SEN]} positions) as
\begin{equation}
H\!\left(\pi_\theta(\cdot \mid \sigma)\right)
= \mathbb{E}\left[-\frac{1}{m} 
\sum_{i = 1}^{m} \sum_{a_i \in \{0,1\}}
\pi_\theta(a_i \mid \sigma)\,\log \pi_\theta(a_i \mid \sigma)\right]
\end{equation}
The final training loss is 
\begin{equation}
\mathcal{L}_{\text{total}}(\theta) = \,\mathcal{L}_{\text{RL}}(\theta)\;-\;\lambda\,H(\pi_\theta(\cdot \mid \sigma)),
\label{eq:total_loss}
\end{equation}
where $\lambda$ controls the trade-off between exploitation and exploration. In practice, we set $\lambda\geq0.1$, which encourages more exploration.

\subsection{Best Trial Memory Update}
\begin{algorithm}[H]
\caption{\met Update the best trial memory}
\label{alg:memory}
\KwIn{
    Buffer $C$ (a dictionary mapping each state $\sigma$ to its best trial), state $\sigma$, group actions $\mathbf{A}$, group rewards $R$
}
\KwOut{Updated buffer $C$}

\For{$(A_i, r_i)$ $\in$ $(\mathbf{A}, R)$}{
  \uIf{$r_i > 0$ and ($\sigma \notin C$ or $r_i \geq C[\sigma].\mathrm{reward}$)}{
    $C[\sigma] \leftarrow (A_i,r_i)$
  }
}
\Return $C$
\end{algorithm}

\subsection{Finetuning the Sentence Selector}
\begin{algorithm}[H]
\caption{\met Hybrid Reinforcement \& Supervised Learning}
\label{alg:training}
\KwIn{Training data $\Delta$, TargetLLM, Intermediate trained policy $\pi_\theta$}
\textbf{Hyperparameters:} Adam optimizer (lr $=1{\times}10^{-6}$), group size $g=8$, $\alpha=0.95$, epochs $=5$, entropy coefficient $\lambda=0.1$, clip value $\epsilon=0.2$, rollout size $=128$

\KwOut{Optimized policy $\pi_\theta^\star$}

Initialize the rollout buffer $\mathcal{B} \leftarrow \emptyset$ and the best trial memory $C\leftarrow \emptyset$\;

\For{example $(q, y_q, S_q) \in \Delta$}{
  $\sigma \leftarrow I(q, S_q)$\;
  Get actions $A^\star \leftarrow \arg\max_{A \in \{0,1\}^m} \pi_\theta(\cdot \mid \sigma)$\;
  Build compressed context $\mathbb{C} \leftarrow \mathrm{Select}(S_q, A^\star)$\;
  Obtain output $y \leftarrow \mathrm{TargetLLM}(q, \mathbb{C})$\;
  Compute reward $r \leftarrow \mathrm{Reward}(y_q, y, S_q, \mathbb{C})$\;
  UpdateMemory$(C, \sigma, \langle A^\star \rangle, \langle r \rangle)$ using Algorithm~\ref{alg:memory}\;
}

reference model $\pi_{ref} \leftarrow \pi_\theta$\;
\For{epoch $=1$ \KwTo $\,\mathrm{epochs}$}{
   reference model $\pi_{ref} \leftarrow \pi_\theta$\;
  \For{example $(q, y_q, S_q) \in \Delta$}{
    $\sigma \leftarrow I(q, S_q)$\;
    Sample actions $\mathbf{A} \leftarrow \{A_i\}_{i=1}^g \sim \pi_\theta(\cdot \mid \sigma)$ (Eq.~\ref{eq:sampling-categorical})\;
    \For{$i=1$ \KwTo $g$}{
      $\mathbb{C}_i \leftarrow \mathrm{Select}(S_q, A_i)$\;
      $y_i \leftarrow \mathrm{TargetLLM}(q, \mathbb{C}_i)$\;
      $r_i \leftarrow \mathrm{Reward}(y_q, y_i, S_q, \mathbb{C}_i)$\;
    }
    Let $R \leftarrow \{r_i\}_{i=1}^g$\;
    UpdateMemory$(C, \sigma, \mathbf{A}, R)$ using Algorithm~\ref{alg:memory}\;
    Compute relative advantages $Z$ from $R$\;
    Append $(\sigma, \mathbf{A}, Z)$ to the rollout collection $\mathcal{B}$\;

    \If{$|\mathcal{B}| \ge \text{rollout size}$}{
      Update $\pi_\theta$ with final training loss (Eq.~\ref{eq:total_loss})\;
      reference model $\pi_{ref} \leftarrow \pi_\theta$\;
      Reset $\mathcal{B} \leftarrow \emptyset$\;
    }
  }

  \For{each $(S_b, A_b) \in C$}{
    Update $\pi_\theta$ with supervised cross-entropy on $(S_b, A_b)$\;
  }
}

\Return $\pi_\theta^\star$\;
\end{algorithm}

\newpage
\twocolumn

\section{Additional Implementation Details}
\label{sec:additional-implementation-details}

\subsection{Input Representation of the Sentence Selector}

\begin{figure*}[h]
    \centering
    \includegraphics[width=1.0\linewidth]{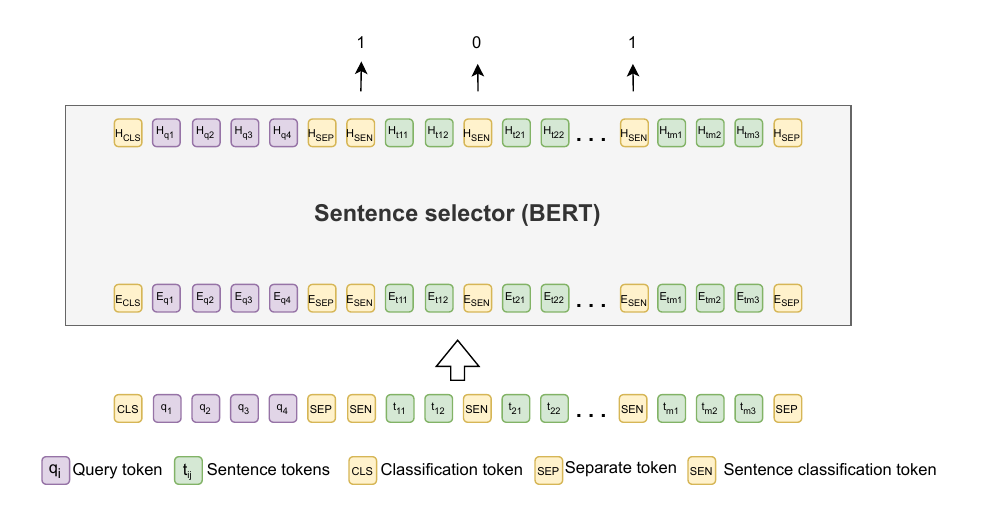}
    \caption{Input and output of the sentence selector. The input begins with the \texttt{[CLS]} token and the query $q$, then followed by the candidate sentences. Each sentence is prepended with a special \texttt{[SEN]} token. 
    The \texttt{[SEP]} tokens were inserted after the query $q$ and after the last token of the last candidate sentence.
    The sentence selector produces token-level outputs, but only predictions at the \texttt{[SEN]} positions are used to determine whether each corresponding sentence is retained or discarded.}
    \label{fig:model-input}
\end{figure*}

For the sentence selector, we use ModernBERT-base~\citep{modernbert} ($\sim$149M parameters) whose input representation is illustrated in Figure~\ref{fig:model-input}.

\subsection{Training Configuration}
\label{sec:training-configuration}
Intermediate finetuning of the sentence selector was conducted using a GeForce RTX 3090 GPU, while final finetuning and inference were conducted using an NVIDIA A100-SXM4-40GB GPU.

\paragraph{Intermediate Finetuning.}  
\begin{itemize}
    \setlength{\itemsep}{0pt}
    \item Learning rate: $2 \times 10^{-5}$
    \item Number of epochs: 2
    \item Weight decay: 0.01
    \item Batch size: 4
    \item Precision: bfloat16
    \item Optimizer: AdamW
    \item Loss: cross-entropy with balanced class weights
\end{itemize}

\paragraph{Final Finetuning.}  
\begin{itemize}
    \setlength{\itemsep}{0pt}
    \item \textbf{Model Setup}
    \begin{itemize}
        \setlength{\itemsep}{0pt}
        \item Target model: Gemma-2-9B-IT
        \item Quantization: BitsAndBytes 4-bit (nf4), compute dtype: fp16
        \item Sampling: greedy
        \item Model maximum length: 2048
    \end{itemize}
    \item Learning rate: $1 \times 10^{-6}$
    \item Number of epochs: 5
    \item Optimizer: AdamW
    \item Precision: bfloat16
    \item Rollout size: 128
    \item Group size: 8
    \item Backward batch size: 8
    \item Updates per iteration: 4
    \item Entropy coefficient $\lambda: 0.1$
    \item Max gradient norm: 0.5
    \item Clip value $\epsilon: 0.2$
    \item Target KL: 0.02
    \item Reward weight ($\alpha$): 0.95
    \item Supervised loss: cross-entropy with balanced class weights
\end{itemize}

Note that we selected the best intermediate finetuning model checkpoint and the best finetuned model checkpoint based on the validation loss and the validation reward observed, respectively.

\begin{table*}[t]
\centering
\footnotesize
\setlength{\tabcolsep}{8pt}
\renewcommand{\arraystretch}{1.2}

\begin{tabular}{l|c|c|c}
\hline
\textbf{Dataset} & \textbf{Train} & \textbf{Validation} & \textbf{Test} \\
\hline
\multicolumn{4}{l}{\textit{Base datasets}} \\
\hline
HotpotQA & 90,447 & -- & 7,405 \\
2WikiMultiHopQA & 167,454 & -- & 12,576 \\
Natural Questions & 152,148 & -- & 5,499 \\
TriviaQA & 138,384 & -- & 17,944 \\
\hline
\multicolumn{4}{l}{\textit{Intermediate finetuning}} \\
\hline
HotpotQA & 75,059 & 8,340 & -- \\
2WikiMultiHopQA & 92,595 & 10,289 & -- \\
Natural Questions & 133,060 & 14,785 & -- \\
TriviaQA & 123,670 & 13,742 & -- \\
\hline
\multicolumn{4}{l}{\textit{Final finetuning}} \\
\hline
HotpotQA & 19,207 & 1,000 & -- \\
2WikiMultiHopQA & 15,522 & 1,000 & -- \\
Natural Questions & 18,687 & 1,000 & -- \\
TriviaQA & 21,136 & 1,000 & -- \\
\hline
\end{tabular}

\caption{Number of examples in each dataset before and after filtering. The base dataset refers to the raw data, while the intermediate finetuning and final finetuning stages indicate the dataset sizes used in each phase.}
\label{tab:datasets}
\end{table*}

\subsection{Dataset Information}

We trained and evaluated \met on four English open-domain QA datasets, including two multi-document benchmarks (HQA and 2WikiQA) and two single-document benchmarks (NQ and TQA). These datasets were used in both intermediate finetuning and final finetuning. Dataset statistics are summarized in Table~\ref{tab:datasets}.

For preprocessing, we used the NLTK 3.9.1~\citep{bird-loper-2004-nltk} sentence tokenizer to segment documents into sentences.

During intermediate finetuning, we excluded two types of examples: (1) examples with $|S| > 5120$ tokens, which rarely occur due to the small $k$ used in document reranking, and (2) samples with ground-truth answers such as “yes”, “no”, or “no-answer”, since they do not correspond to concrete evidence in the retrieved text and may introduce misleading supervision.

During final finetuning, we excluded samples exceeding 2048 tokens due to hardware memory constraints.

\subsection{LLM Prompt Template}

For both Gemma-2-9B-IT and Llama-3.1-8B-Instruct, we adopted the prompt template by~\citet{jung2024familiarityawareevidencecompressionretrievalaugmented} to query the LLMs:

\begin{mdframed}
{
\fontfamily{cmss}\selectfont
You are an expert in Question Answering. Your job is to answer questions in 1 to 5 words based on the given context. \\
Question: \{question\} \\
Context: \{context\} \\
Answer:
}
\end{mdframed}

\subsection{Inference Configuration for Target LLMs}
All the results are reported from a single run. The configurations of the target LLMs are shown below.

\paragraph{Gemma-2-9B-IT.}  
\begin{itemize}
    \setlength{\itemsep}{0pt}
    \item Sampling: greedy
    \item Quantization: BitsAndBytes 4-bit (nf4), dtype: float16
    \item Max new tokens: 10
    \item Model maximum length: 6144
\end{itemize}

\paragraph{Llama-3.1-8B-Instruct.}  
\begin{itemize}
    \setlength{\itemsep}{0pt}
    \item Sampling: greedy
    \item Quantization: BitsAndBytes 4-bit (fp4), dtype: float32
    \item Max new tokens: 10
    \item Model maximum length: 5120
\end{itemize}

\subsection{Baseline Details}
\label{sec:baseline-detail}

This section provides additional details about the baselines in our experiments.
\begin{itemize}
    \item \textbf{Raw Documents}: The top-30 retrieved documents are used directly without any compression.  
    \item \textbf{Reranker-only methods}: The top-30 retrieved documents are reordered using a document reranker and the top-10 documents are selected as the compressed context. We evaluate three rerankers in this setting: BGE ($\sim$278M parameters), Jina ($\sim$278M parameters), and GTE ($\sim$149M parameters).
    \item \textbf{LongLLMLingua} \citep{jiang-etal-2024-longllmlingua}: A task-aware method built on Llama-2-7B\footnote{\url{https://huggingface.co/NousResearch/Llama-2-7b-hf}} that ranks documents and selects tokens using contrastive perplexity scoring. We set the dynamic compression rate fixed at 0.3, with the rate set to 0.45 for the $\sim$5× compression.
    \item \textbf{LLMLingua2} \citep{pan-etal-2024-llmlingua}: A task-agnostic compression method trained via label distillation from GPT-4 on the MeetingBank dataset~\citep{hu-etal-2023-meetingbank}. We use the XLM-RoBERTa-large model\footnote{https://huggingface.co/microsoft/llmlingua-2-xlm-roberta-large-meetingbank} ($\sim$550M parameters) ~\citep{conneau2020unsupervisedcrosslingualrepresentationlearning}, which follows the LLMLingua2 training recipe. The compression rate is set to 0.2.
    \item \textbf{Recomp-Extractive} \citep{xu2024recomp}: An extractive approach that assigns scores at the sentence level using a dual-encoder model. We adopt the Contriever models ($\sim$110M parameters) released by the authors for each dataset. For 2WikiQA, where no finetuned models are available, we use the Contriever finetuned on HQA. Following the previous research~\citep{yoon-etal-2024-compact} and similar compression ratio with our method, the top four sentences are selected as context.
    \item \textbf{Recomp-Abstractive} \citep{xu2024recomp}: An abstractive approach based on T5-large ($\sim$770M) trained with summarization distillation over NQ, TQA, and HQA. The model compresses the retrieved evidence into shorter summaries. For 2WikiQA and HQA, we use the -trained version, while for TQA and NQ, dataset-specific models are applied.  
    \item \textbf{CompAct} \citep{yoon-etal-2024-compact}: An iterative abstractive compression method implemented with Mistral-7B-Instruct\footnote{https://huggingface.co/cwyoon99/CompAct-7b} \citep{jiang2023mistral7b}. The model repeatedly summarizes the input until the target model signals completion. The number of documents per segment is set to 5, following the original implementation.
    \item \textbf{Context-aware Prompt Compression (CPC)} \citep{Liskavets_Ushakov_Roy_Klibanov_Etemad_Luke_2025}: A context-aware compression approach that assigns relative importance scores to sentences conditioned on the given question. The model is trained on a synthetic dataset using a contrastive learning objective and implemented an adaptor with Llama-3.2-1B-Instruct\footnote{deadcode99/cpc-1.0-llama-1b-ds-v5-iter66-lora-bidirectional-attn}~\citep{jiang2023mistral7b}. We follow the original paper and set the target compressed length to 900 tokens, which corresponds to approximately a 5x compression when using 30 documents from the retriever.
    \item \textbf{DAC} \citep{zhao-etal-2025-dac}: A prompt-agnostic compression method that integrates entropy and attention signals to compute token-level importance scores. We use the Qwen2-0.5B-Instruct \footnote{https://huggingface.co/Qwen/Qwen2-0.5B-Instruct} model~\citep{yang2024qwen2technicalreport} with the default configuration, and compression ratio set to 0.825 (around 5x).
\end{itemize}

\section{Additional Results and Analyses}
\label{sec:appendix-experiments}

\subsection{Examples}
To illustrate, we show a successful case of \met in Table~\ref{tab:compression-example}. In this case, Recomp-Abstractive and CompAct generated compressed contexts that are factually incorrect or irrelevant to the question, leading to wrong LLM predictions. By contrast, LongLLMLingua and \met remained correctly grounded in the retrieved contexts, but \met used 8.6 times fewer tokens than LongLLMLingua (88 vs. 758 tokens).

\begin{table*}[h]
\centering
\footnotesize
\begin{tabular}{p{1.5cm} c c p{8.05cm} p{2.4cm}}
\toprule
\multicolumn{5}{l}{\textbf{Question} (from HQA): Stein had a part in the 1994 superhero fantasy film directed by whom} \\
\multicolumn{5}{l}{\textbf{Ground-truth answer}: Charles Russell }\\
\midrule
\textbf{Method} & \textbf{$1/\tau$} $\uparrow$ & $|\mathbb{C}|$ $\downarrow$ &\textbf{Compressed Context} & \textbf{LLM Prediction} \\
\midrule
Long-LLMLingua & 5.01x & 758 & The Mask (film) The Mask is a 1994 American fantasy superhero comedy film directed by Charles Russell, produced by Bob Engelman [...] Ben Stein Benjamin Jeremy (born November 25, 1944) is an American writer, [...] He is well- screen econom teacher "Ferris Bueller's Day Off" (1986) and as Dr. Arthur Neuman in "The Mask" (1994)Son [...]
& Charles Russell \cmark \\
\midrule
Recomp-Abstractive & 189.75x & 20 & 
Ben Stein had a part in the 1994 superhero fantasy film "The Shadow" directed by Russell Mulcahy.
& Russell Mulcahy \xmark \\
\midrule
CompAct & 49.93x & 76 &
The Shadow is a 1994 American superhero film [...] directed by Russell Mulcahy, that stars Alec Baldwin. The film co-stars John Lone, Penelope Ann Miller, 
[...]
It was released to theaters on July 1, 1994 and received generally mixed reviews.
& Russell Mulcahy \xmark \\
\midrule
\met (BGE) & 43.13x & 88 &
He is most well-known on screen as the economics teacher in "Ferris Bueller's Day Off" (1986) and as Dr. Arthur Neuman in "The Mask" (1994) and "Son of
The Mask (film) The Mask is a 1994 American fantasy superhero comedy film directed by Charles Russell, produced by Bob Engelman [...]
& Charles Russell \cmark \\
\bottomrule
\end{tabular}
\caption{
Examples of compressed contexts and predictions from LongLLMLingua, Recomp-Abstrctive, CompAct, and \met (with BGE reranker).
$|\mathbb{C}|$ denotes the number of tokens of the compressed context.
\cmark\ indicates a correct prediction, while \xmark\ indicates an incorrect prediction.}
\label{tab:compression-example}
\end{table*}

\subsection{Performance under the In-Distribution Training Setting} \label{subsec:appendix-out-domain}
We report the results of in-distribution and out-of-distribution testing for the other rerankers and the other target LLM (Llama-3.1-8B-Instruct) in Table~\ref{tab:more-generalize-dataset-result}.
Training on the target dataset yielded both higher compression ratios and stronger QA metrics, while applying the model out of distribution resulted in certain reductions in compression ratios but only minor reductions in the QA metric. 

\begin{table*}[h]
\centering
\footnotesize
\setlength{\tabcolsep}{3pt}
\renewcommand{\arraystretch}{1.25}

\begin{tabular}{l|rccc|rccc|rccc}
\hline
\textbf{Method}
& \multicolumn{4}{c|}{\textbf{2WikiQA}}
& \multicolumn{4}{c|}{\textbf{NQ}}
& \multicolumn{4}{c}{\textbf{TQA}} \\
\cline{2-13}
& $1/\tau$ & Acc & $\mathrm{C}_{0.5}$ & $\mathrm{C}_{1}$
& $1/\tau$ & Acc & $\mathrm{C}_{0.5}$ & $\mathrm{C}_{1}$
& $1/\tau$ & Acc & $\mathrm{C}_{0.5}$ & $\mathrm{C}_{1}$\\
\hline
\multicolumn{13}{l}{\textbf{Gemma-2-9B-IT}} \\
\hline
No reranker (in-distribution)  
& \textbf{16.1x} & \textbf{29.5} & \textbf{0.34} & \textbf{0.45}
& \textbf{23.6x} & \textbf{31.7} & \textbf{0.37} & \textbf{0.48}
& \textbf{48.4x} & \textbf{78.9} & \textbf{0.82} & \textbf{0.87}\\
No reranker (out-of-distribution) 
& 11.7x & 28.9 & \textbf{0.34} & 0.44
& 9.8x & 31.1 & 0.36 & 0.46
& 32.8x & \textbf{78.9} & \textbf{0.82} & \textbf{0.87}\\
\hline
BGE reranker (in-distribution) 
& \textbf{30.3x} & \textbf{25.4} & \textbf{0.30} & \textbf{0.40}
& \textbf{36.4x} & \textbf{29.5} & \textbf{0.34} & \textbf{0.45}
& \textbf{59.5x} & \textbf{80.1} & \textbf{0.83} & \textbf{0.88}\\
BGE reranker (out-of-distribution) 
& 19.4x & 25.0 & 0.29 & \textbf{0.40}
& 14.1x & 29.0 & \textbf{0.34} & 0.44
& 49.6x & 79.5 & \textbf{0.83} & \textbf{0.88}\\
\hline
Jina reranker (in-distribution) 
& \textbf{29.4x} & 25.5 & \textbf{0.30} & 0.40
& \textbf{42.0x} & \textbf{30.0} & \textbf{0.35} & \textbf{0.46}
& \textbf{78.4x} & \textbf{80.1} & \textbf{0.83} & \textbf{0.88}\\
Jina reranker (out-of-distribution) 
& 19.3x & \textbf{25.8} & \textbf{0.30} & \textbf{0.41}
& 18.8x & 29.7 & 0.34 & 0.45
& 50.0x & 79.6 & \textbf{0.83} & \textbf{0.88}\\
\hline
GTE reranker (in-distribution) 
& \textbf{29.4x} & \textbf{25.4} & \textbf{0.30} & \textbf{0.40}
& \textbf{42.3x} & \textbf{30.4} & 
\textbf{0.35} & \textbf{0.46}
& \textbf{64.3x} & \textbf{80.5} & \textbf{0.84} & \textbf{0.89}\\
GTE reranker (out-of-distribution) 
& 20.4x & 25.0 & 0.29 & \textbf{0.40}
& 19.2x & 29.7 & 0.34 & 0.45
& 54.1x & 79.8 & 0.83 & 0.88\\
\hline
\multicolumn{13}{l}{\textbf{Llama-3.1-8B-Instruct}} \\
\hline
No reranker (in-distribution)  
& \textbf{16.1x} & \textbf{21.3} & \textbf{0.25} & \textbf{0.35}
& \textbf{23.6x} & \textbf{27.9} & \textbf{0.33} & \textbf{0.43}
& \textbf{48.4x} & 73.5 & 0.77 & \textbf{0.84}\\
No reranker (out-of-distribution) 
& 11.7x & 20.8 & \textbf{0.25} & 0.34
& 9.8x & 26.8 & 0.31 & 0.41
& 32.8x & \textbf{74.3} & \textbf{0.78} & \textbf{0.84}\\
\hline
BGE reranker (in-distribution) 
& \textbf{30.3x} & \textbf{21.5} & \textbf{0.26} & \textbf{0.35}
& \textbf{36.4x} & \textbf{28.1} & \textbf{0.33} & \textbf{0.44}
& \textbf{59.5x} & \textbf{76.6} & \textbf{0.80} & \textbf{0.86}\\
BGE reranker (out-of-distribution) 
& 19.4x & 20.3 & 0.24 & 0.33
& 14.1x & 27.5 & 0.32 & 0.42
& 49.6x & 76.1 & \textbf{0.80} & \textbf{0.86}\\
\hline
Jina reranker (in-distribution) 
& \textbf{29.4x} & \textbf{21.8} & \textbf{0.26} & \textbf{0.36}
& \textbf{42.0x} & \textbf{28.9} & \textbf{0.34} & \textbf{0.45}
& \textbf{78.4x} & \textbf{77.1} & \textbf{0.81} & \textbf{0.87}\\
Jina reranker (out-of-distribution) 
& 19.3x & 21.3 & 0.25 & 0.35
& 18.8x & 28.2 & 0.33 & 0.44
& 50.0x & 76.2 & 0.80 & 0.86\\
\hline
GTE reranker (in-distribution) 
& \textbf{29.4x} & \textbf{21.6} & \textbf{0.26} & \textbf{0.35}
& \textbf{42.3x} & \textbf{29.6} & \textbf{0.34} & \textbf{0.45}
& \textbf{64.3x} & \textbf{77.3} & \textbf{0.81} & \textbf{0.87}\\
GTE reranker (out-of-distribution) 
& 20.4x & 20.8 & 0.25 & 0.34
& 19.2x & 28.3 & 0.33 & 0.44
& 54.1x & 77.2 & \textbf{0.81} & 0.86\\
\hline
\end{tabular}

\caption{Cross-dataset generalization of \met using Gemma-2-9B-IT and Llama-3.1-8B-Instruct as target models, evaluated both with and without the document reranker.
We report the mean compression ratio ($1/\tau$), accuracy (Acc), and $\mathrm{C}_{\beta}$. The in-distribution rows correspond to training on each dataset individually, while the out-of-distribution rows correspond to training only on HQA and directly applying the model to other datasets. For all of these metrics, higher values are better. Bold numbers indicate the best performance in each setting. }
\label{tab:more-generalize-dataset-result}
\end{table*}

\subsection{Generalization to Another Target Model} \label{subsec:appendix-llama}
We evaluated our method, trained on the HQA dataset with Gemma-2-9B-IT as the target model during training, against baseline approaches under the setting where compressed contexts were provided to another target model which is Llama-3.1-8B-Instruct. Table~\ref{tab:qa-result-llama} reports the results on HQA, 2WikiQA, NQ, and TQA. Overall, the performance trends remain consistent across datasets, with our method achieving the best QA results. Furthermore, \met with a reranker generally results in a higher compression rate, showing the importance of the reranker to our method.

\begin{table*}[h]
\centering
\footnotesize
\setlength{\tabcolsep}{3pt}
\renewcommand{\arraystretch}{1.25}

\begin{tabular}{l|rccc|rccc|rccc|rccc}
\hline
\textbf{Method}
& \multicolumn{4}{c|}{\textbf{HQA}}
& \multicolumn{4}{c|}{\textbf{2WikiQA}}
& \multicolumn{4}{c|}{\textbf{NQ}}
& \multicolumn{4}{c}{\textbf{TQA}} \\
\cline{2-17}
& $1/\tau$ & Acc & $\mathrm{C}_{0.5}$ & $\mathrm{C}_{1}$
& $1/\tau$ & Acc & $\mathrm{C}_{0.5}$ & $\mathrm{C}_{1}$
& $1/\tau$ & Acc & $\mathrm{C}_{0.5}$ & $\mathrm{C}_{1}$
& $1/\tau$ & Acc & $\mathrm{C}_{0.5}$ & $\mathrm{C}_{1}$ \\
\hline
Raw Documents
& 1.0x & 24.9 & 0.00 & 0.00
& 1.0x & 16.7 & 0.00 & 0.00
& 1.0x & 25.3 & 0.00 & 0.00
& 1.0x & 69.7 & 0.00 & 0.00 \\
\hline
\multicolumn{17}{l}{\textit{Reranker-only methods}} \\
BGE & 3.0x & 29.6  & 0.33 & 0.41 & 3.0x & 19.2 & 0.22 & 0.30 & 3.0x & 26.6 & 0.30 & 0.38 & 3.0x & 75.3 & 0.73 & 0.71 \\
Jina & 3.0x & 29.2 & 0.33 & 0.41 & 3.0x & 19.3 & 0.22 & 0.30 & 3.0x & 28.1 & 0.32 & 0.40 & 3.0x & 75.6 & 0.74 & 0.71 \\
GTE & 3.0x & 29.0 & 0.33 & 0.40 & 3.0x & 19.2 & 0.22 & 0.30 & 3.0x & 28.4 & 0.32 & 0.40 & 3.0x & 75.9 & 0.74 & 0.71 \\
\hline
\multicolumn{17}{l}{\textit{Compression frameworks}} \\
LLMLingua2 & 5.4x & 21.4 & 0.25 & 0.34 & 5.5x & 17.3 & 0.21 & 0.29 & 5.3x & 22.1 & 0.26 & 0.35 & 5.4x & 68.8 & 0.71 & 0.75 \\
DAC & 5.1x & 19.9 & 0.23 & 0.32 & 5.0x & 16.6 & 0.20 & 0.27 & 5.2x & 19.4 & 0.23 & 0.31 & 5.2x & 67.5 & 0.70 & 0.74 \\
CPC & 5.6x & 26.7 & 0.31 & 0.40 & 5.9x & 17.7 & 0.21 & 0.29 & 5.2x & 25.8 & 0.30 & 0.39 & 5.4x & 74.2 & 0.76 & 0.78 \\
LongLLMLingua & 5.1x & 27.7 & 0.32 & 0.41 & 5.1x & 18.8 & 0.22 & 0.30 & 5.1x & 24.4 & 0.28 & 0.37 & 5.1x & 73.6 & 0.75 & 0.77 \\
Recomp$_{\mathrm{Extractive}}$ & 29.2x & 24.0 & 0.28 & 0.38 & 29.3x & 16.6 & 0.20 & 0.28 & 29.8x & 22.3 & 0.26 & 0.36 & 32.0x & 69.7 & 0.74 & 0.81 \\
Recomp$_{\mathrm{Abstractive}}$ & \textbf{132.1x} & 27.8 & 0.32 & 0.43 & \textbf{126.9x} & 21.3 & 0.25 & 0.35 & \textbf{48.3x} & 26.3 & 0.31 & 0.41 & \textbf{139x} & 74.3 & 0.78 & 0.85 \\
CompAct & 48.7x & 30.5 & 0.35 & \textbf{0.47} & 54.7x & 18.8 & 0.22 & 0.32 & 44.2x & 27.9 & 0.33 & 0.43 & 50.2x & 75.9 & 0.79 & 0.86 \\
\hline
\multicolumn{17}{l}{\textit{Our work:} \met} \\
No reranker & 26.8x & 29.6 & 0.34 & 0.45 & 11.7x & 20.8 & \textbf{0.25} & 0.34
& 9.8x & 26.8 & 0.31 & 0.41
& 32.8x & 74.3 & 0.78 & 0.84 \\
\hdashline
BGE reranker & 39.5x & \textbf{31.3} & \textbf{0.36} & \textbf{0.47} & 30.3x & 21.5 & 0.25 & 0.35 & 36.4x & 28.1 & 0.33 & 0.44 & 59.5x & 76.6 & 0.80 & 0.86 \\
Jina reranker & 38.5x & 31.0 & \textbf{0.36} & \textbf{0.47} & 29.4x & \textbf{21.8} & \textbf{0.26} & \textbf{0.36} & 42.0x & 28.9 & \textbf{0.34} & \textbf{0.45} & 78.4x & 77.1 & \textbf{0.81} & \textbf{0.87} \\
GTE reranker & 40.4x & 31.0 & \textbf{0.36} & \textbf{0.47} & 29.4x & 21.6 & \textbf{0.26} & 0.35 & 42.3x & \textbf{29.6} & \textbf{0.34} & \textbf{0.45} & 64.3x & \textbf{77.3} &  \textbf{0.81} & \textbf{0.87} \\
\hline
\end{tabular}

\caption{Results on four QA benchmarks.
We report the mean of compression ratio ($1/\tau$), accuracy (Acc), $\mathrm{C}_{\beta}$ for Llama-3.1-8B-Instruct as the target model. For all of these metrics, higher values are better. The best results in each column are marked in bold.
}
\label{tab:qa-result-llama}
\end{table*}

\subsection{Generalization to Long Generation Tasks}
\label{subsec}

\met is designed primarily for RAG-based QA tasks, as reflected by both the supervision used in Phase I and the reward function employed during RL training in Phase II. Consequently, the training objectives are not directly designed for long-form generation tasks such as document summarization. To examine whether \met can nevertheless generalize to such tasks, we apply the \met model trained on HQA directly to three summarization datasets from LongBench: GovReport, Multi-News, and QMSum. The results are reported in Table~\ref{tab:long-generation-result}.

\begin{table*}[h]
\centering
\footnotesize
\setlength{\tabcolsep}{3pt}
\renewcommand{\arraystretch}{1.25}

\setlength{\tabcolsep}{5pt}
\begin{tabular}{l|rc|rc|rc}
\hline
\textbf{Method}
& \multicolumn{2}{c|}{\textbf{GovReport}}
& \multicolumn{2}{c|}{\textbf{Multi-News}}
& \multicolumn{2}{c}{\textbf{QMSum}} \\
\cline{2-7}

& $1/\tau$ & ROUGE-L
& $1/\tau$ & ROUGE-L
& $1/\tau$ & ROUGE-L \\
\hline

Raw documents
& 1.0x & 18.8
& 1.0x & \textbf{18.2}
& 1.0x & \textbf{19.2} \\

LLMLingua2
& 5.2x & 16.8
& \textbf{5.7x} & 16.8
& 5.5x & 17.6 \\

DAC
& 6.0x & 17.4
& 4.9x & 17.1
& \textbf{6.3x} & 17.2 \\

CPC
& 5.1x & 17.2
& 4.7x & 16.3
& 5.2x & 17.7 \\

LongLLMLingua
& 4.6x & \textbf{19.0}
& 3.1x & 18.1
& 3.8x & 19.0 \\

\met~fixed
& 5.0x & 17.4
& 5.3x & 16.9
& 4.9x & 18.1 \\

\met~adaptive
& \textbf{12.3x} & 16.9
& 1.4x & 17.8
& 1.9x & 18.8 \\

\hline
\end{tabular}%

\caption{
Generalization of \met to long-generation tasks from LongBench.
The \met model is trained on HQA and directly applied to GovReport,
Multi-News, and QMSum without task-specific training.
We report the mean compression ratio ($1/\tau$) and ROUGE-L score.
Higher values are better for both metrics.
Bold numbers indicate the best performance in each setting.
}
\label{tab:long-generation-result}
\end{table*}

As expected, \met does not consistently outperform the baselines on these long-generation tasks. Although \met with a fixed compression ratio achieves compression ratios comparable to several existing compression methods, its ROUGE-L scores are generally lower than those of LongLLMLingua. The adaptive variant exhibits different behavior across datasets, achieving a substantially higher compression ratio of 12.3x on GovReport, while selecting considerably lower compression ratios of 1.4x and 1.9x on Multi-News and QMSum, respectively.

Among the evaluated methods, LongLLMLingua performs relatively well in this setting, maintaining strong ROUGE-L scores while reducing the input context. This is reasonable because LongLLMLingua is designed for general generation tasks, whereas \met is specifically optimized for RAG-based QA. Overall, these results suggest that although \met can be directly applied to long-generation tasks without additional training, its advantages are less pronounced outside the QA setting for which it was designed.

\subsection{Additional Experiments with Recent Retrievers}
\label{sec:recent-retrievers}

To investigate whether \met remains beneficial when combined with stronger
retrievers and target LLMs, we conduct additional experiments using recent
retrieval methods and a more capable target LLM. In particular, we examine
whether \met can further improve QA performance even when the underlying
retrieval or generation component is improved.

We consider five retrieval approaches from two directions:

\begin{itemize}
    \item \textbf{Improve embeddings for retrieval.}
    We replace Contriever-MS MARCO, used in our paper, with:
    \begin{itemize}
        \item \textbf{Voyage 4 Lite\footnote{https://huggingface.co/voyageai/voyage-4-lite}}, a commercial
        embedding model.
        \item \textbf{F2LLM-v2-8B}~\cite{zhang2026f2llmv2inclusiveperformantefficient}, strongest open-source
        retriever according to the 2026 model on MTEB benchmark~\cite{muennighoff-etal-2023-mteb}.
    \end{itemize}

    \item \textbf{Improve retrieval pipelines.}
    We consider three approaches that use more sophisticated retrieval
    pipelines:
    \begin{itemize}
        \item \textbf{LLMChainExtractor}~\cite{verma2024contextualcompressionretrievalaugmentedgeneration} with GPT-4.1 Nano, which uses an LLM
        to extract query-relevant passages or sentences from each retrieved
        document.
        \item \textbf{EmbeddingsRedundantFilter}~\cite{verma2024contextualcompressionretrievalaugmentedgeneration} with
        text-embedding-3-large and a similarity threshold of 0.7,
        which removes semantically redundant documents based on their
        embeddings.
        \item \textbf{MultiQueryRetriever}\footnote{https://reference.langchain.com/python/langchain-classic/retrievers/multi_query/MultiQueryRetriever}, which generates multiple query
        reformulations to improve document recall.
    \end{itemize}
\end{itemize}

We evaluate these methods on 1,000 HQA examples using two target LLMs,
Gemma-2-9B-IT~\cite{gemma_2024} and Claude Sonnet 4.5\footnote{https://platform.claude.com/docs/en/models/sonnet-4-5/overview}. The results are shown in
Table~\ref{tab:recent-retrievers}. The setting marked with * corresponds to
the retrieval configuration mainly used throughout our paper. For LLMChainExtractor, the retrieved documents are already highly compressed; therefore, we do not apply \met to avoid further compression.

\begin{table*}[h]
\centering
\footnotesize
\setlength{\tabcolsep}{4pt}
\renewcommand{\arraystretch}{1.15}

\begin{tabular}{l|c|c|rccc}
\hline
\textbf{Retriever}
& \textbf{Step 1 (BGE)}
& \textbf{Step 2}
& \textbf{$1/\tau$}
& \textbf{Acc}
& $\mathbf{C}_{0.5}$
& $\mathbf{C}_{1}$ \\
\hline

\multicolumn{7}{l}{\textbf{Gemma-2-9B-IT}} \\
\hline

Contriever-MS MARCO (Top 30)$^{*}$
& Top 10 & \multicolumn{1}{c|}{\cmark}
& 41.1x & 35.7 & 0.41 & 0.52 \\

Voyage 4 Lite (Top 6)
& -- & --
& 5.0x & 29.3 & 0.34 & 0.43 \\

Voyage 4 Lite (Top 6)
& -- & \multicolumn{1}{c|}{\cmark}
& 37.3x & 30.2 & 0.35 & 0.46 \\

Voyage 4 Lite (Top 30)
& Top 6 & \multicolumn{1}{c|}{\cmark}
& 51.6x & 35.3 & 0.40 & 0.52 \\

F2LLM-v2-8B (Top 6)
& -- & --
& 5.0x & 32.6 & 0.37 & 0.46 \\

F2LLM-v2-8B (Top 6)
& -- & \multicolumn{1}{c|}{\cmark}
& 40.6x & 32.9 & 0.38 & 0.49 \\

F2LLM-v2-8B (Top 30)
& Top 6 & \multicolumn{1}{c|}{\cmark}
& 50.4x & \textbf{36.9} & \textbf{0.42} & \textbf{0.54} \\

Contriever-MS MARCO (Top 30) + LLMChainExtractor
& -- & --
& 14.5x & 31.5 & 0.36 & 0.47 \\

Contriever-MS MARCO (Top 30) + EmbeddingsRedundantFilter
& -- & --
& 1.6x & 28.9 & 0.30 & 0.33 \\

Contriever-MS MARCO (Top 30) + EmbeddingsRedundantFilter
& Top 10 & \multicolumn{1}{c|}{\cmark}
& 34.0x & 29.5 & 0.34 & 0.45 \\

Voyage 4 Lite + MultiQueryRetriever (Top 10)
& -- & --
& 6.1x & 28.6 & 0.33 & 0.43 \\

Voyage 4 Lite + MultiQueryRetriever (Top 10)
& -- & \multicolumn{1}{c|}{\cmark}
& 50.9x & 30.0 & 0.35 & 0.46 \\

Voyage 4 Lite + MultiQueryRetriever (Dynamic)
& Top 10 & \multicolumn{1}{c|}{\cmark}
& \textbf{90.7x} & \textbf{36.9} & \textbf{0.42} & \textbf{0.54} \\

\hline
\multicolumn{7}{l}{\textbf{Claude Sonnet 4.5}} \\
\hline

Contriever-MS MARCO (Top 30)$^{*}$
& Top 10 & \multicolumn{1}{c|}{\cmark}
& 41.1x & 43.6 & 0.49 & 0.60 \\

Voyage 4 Lite (Top 6)
& -- & --
& 5.0x & 40.8 & 0.45 & 0.54 \\

Voyage 4 Lite (Top 6)
& -- & \multicolumn{1}{c|}{\cmark}
& 37.3x & 39.3 & 0.45 & 0.56 \\

Voyage 4 Lite (Top 30)
& Top 6 & \multicolumn{1}{c|}{\cmark}
& 51.6x & 44.0 & 0.49 & 0.61 \\

F2LLM-v2-8B (Top 6)
& -- & --
& 5.0x & 43.4 & 0.48 & 0.56 \\

F2LLM-v2-8B (Top 6)
& -- & \multicolumn{1}{c|}{\cmark}
& 40.6x & 43.0 & 0.48 & 0.60 \\

F2LLM-v2-8B (Top 30)
& Top 6 & \multicolumn{1}{c|}{\cmark}
& 50.4x & \textbf{47.0} & \textbf{0.52} & \textbf{0.64} \\

Contriever-MS MARCO (Top 30) + LLMChainExtractor
& -- & --
& 14.5x & 43.0 & 0.48 & 0.59 \\

Contriever-MS MARCO (Top 30) + EmbeddingsRedundantFilter
& -- & --
& 1.6x & 24.4 & 0.26 & 0.30 \\

Contriever-MS MARCO (Top 30) + EmbeddingsRedundantFilter
& Top 10 & \multicolumn{1}{c|}{\cmark}
& 34.0x & 38.4 & 0.44 & 0.55 \\

Voyage 4 Lite + MultiQueryRetriever (Top 10)
& -- & --
& 6.1x & 35.1 & 0.40 & 0.49 \\

Voyage 4 Lite + MultiQueryRetriever (Top 10)
& -- & \multicolumn{1}{c|}{\cmark}
& 50.9x & 37.8 & 0.43 & 0.55 \\

Voyage 4 Lite + MultiQueryRetriever (Dynamic)
& Top 10 & \multicolumn{1}{c|}{\cmark}
& \textbf{90.7x} & 44.3 & 0.50 & 0.61 \\

\hline
\end{tabular}

\caption{
Performance of recent retrieval methods with and without \met on
1,000 HQA examples. Step~1 denotes BGE reranking, where ``Top $k$''
indicates the number of documents retained by the reranker. Step~2 denotes
the sentence selector, where \cmark indicates that sentence selection is
applied. $^{*}$ denotes the retrieval configuration mainly used throughout
our paper.
}
\label{tab:recent-retrievers}
\end{table*}

The results in Table~\ref{tab:recent-retrievers} show that \met remains effective with recent retrieval methods.
First, \met with the standard retriever used in our paper outperforms all
advanced retrievers without \met. For example, with Gemma-2-9B-IT, the
standard setting (Contriever-MS MARCO) achieves 35.7 accuracy, compared with 32.6 for
F2LLM-v2-8B and 29.3 for Voyage 4 Lite without \met. With Claude Sonnet 4.5,
the corresponding accuracy is 43.6, compared with 43.4 and 40.8,
respectively. Second, \met provides substantial gains even when combined with stronger
retrieval methods. For example, F2LLM-v2-8B improves from 32.6 to 36.9 with
\met for Gemma-2-9B-IT, and from 43.4 to 47.0 for Claude Sonnet 4.5.
Similarly, Voyage 4 Lite with MultiQueryRetriever improves from 28.6 to
36.9 for Gemma-2-9B-IT and from 35.1 to 44.3 for Claude Sonnet 4.5 when
combined with the full \met pipeline. These results demonstrate that the
benefits of \met are complementary to improvements in the retrieval pipeline
rather than being limited to a particular retriever.

\subsection{Hallucinations in Compressed Contexts}

As discussed in the Introduction, we hypothesize that \met produces fewer hallucinations than abstractive methods due to its extractive nature. To evaluate this, we use natural language inference (NLI) on the HQA test set.
For each example, the original context is treated as the premise and the compressed context as the hypothesis. We then obtain NLI predictions from three commercial LLMs (Gemini 2.5 Flash Lite\footnote{https://ai.google.dev/gemini-api/docs/models/gemini-2.5-flash-lite}, GPT-4.1 nano\footnote{https://developers.openai.com/api/docs/models/gpt-4.1-nano}, and Claude 3 Haiku~\cite{TheC3}).
Table~\ref{tab:nli_full} reports the number of entailment, neutral, and contradiction predictions where we treat both neutral and contradiction predictions as hallucinations, since neither category is fully supported by the reference context. Across all models, \met (BGE reranker) consistently yields more entailment cases and fewer contradiction cases than both CompAct and Recomp-Abstractive, indicating improved faithfulness.
\met achieves an average hallucination rate of 25\%, compared to 34\% for CompAct and 53\% for Recomp-Abstractive. These results confirm that \met introduces substantially fewer hallucinations than abstractive approaches.

\begin{table*}[h]
\centering
\footnotesize
\setlength{\tabcolsep}{4pt}
\renewcommand{\arraystretch}{1.2}

\begin{tabular}{l|l|c|c|c}
\hline
\textbf{Model} & \textbf{Method} & \textbf{Entailment} & \textbf{Neutral} & \textbf{Contradiction} \\
\hline
 & CompAct & 81.2\% & 15.8\% & 3.0\% \\
Gemini 2.5 Flash Lite & Recomp-Abstractive & 45.7\% & 35.4\% & 18.9\% \\
 & \met (BGE) & \textbf{86.0\%} & \textbf{11.4\%} & \textbf{2.6\%} \\
\hline
 & CompAct & 50.6\% & 1.3\% & 48.1\% \\
GPT-4.1 nano & Recomp-Abstractive & 43.7\% & 0.6\% & 55.8\% \\
 & \met (BGE) & \textbf{61.8\%} & \textbf{0.2\%} & \textbf{38.0\%} \\
\hline
 & CompAct & 66.7\% & 24.8\% & 8.5\% \\
Claude 3 Haiku & Recomp-Abstractive & 50.3\% & 21.0\% & 28.7\% \\
 & \met (BGE) & \textbf{77.1\%} & \textbf{18.7\%} & \textbf{4.2\%} \\
\hline
\end{tabular}

\caption{NLI label counts on HQA (7,405 examples). Higher entailment and lower neutral/contradiction indicate better faithfulness and fewer hallucinations.}
\label{tab:nli_full}
\end{table*}

\subsection{Inference Latency And Computational Cost}  
Table~\ref{tab:efficiency-lat} reports the compression latency, the target LLM reading latency, and the total latency, all averaged from 500 HQA examples with a batch size of 1.
\met substantially reduces LLM reading time (up to 55.4\%), while introducing only a small compression overhead compared to no compression. As a result, it reduces total latency up to a 53.3\% when using the Jina reranker.
\met is among the fastest methods, together with Recomp-Extractive

To further evaluate computational efficiency, we measure the computational cost of compression in terms of Teraflops (TFLOPS) under the same evaluation setting (HQA, 500 samples, batch size = 1), as shown in Table~\ref{tab:efficiency-tflops}. Since \met employs a lightweight sentence-level compressor, it incurs substantially lower computational cost than most existing compression frameworks. In particular, \met with the Jina reranker requires only 1.2 TFLOPs, consisting of 0.9 TFLOPs for reranking and 0.3 TFLOPs for sentence-level compression, which is the lowest compression cost among the compared methods. These results demonstrate that \met achieves both low inference latency and low computational cost, making it suitable for practical, user-facing applications.

\begin{table}[h]
\centering
\footnotesize
\setlength{\tabcolsep}{6pt}
\renewcommand{\arraystretch}{1.2}

\begin{tabular}{l|c|c|c}
\hline
\textbf{Method} 
& \textbf{Comp (s)} 
& \textbf{Read (s)} 
& \textbf{Total (s)} \\
\hline
Raw documents & -- & 1.84 & 1.84\\
\hline 

LLMLingua2 & 0.28 & 0.84 & 1.12 \\
DAC & 1.46 & 0.79 & 2.25 \\
CPC & 0.26 & 0.81 & 1.07 \\
LongLLMLingua & 2.31 & 0.90 & 3.21 \\
Recomp-Extractive & 0.19 & 0.74 & 0.93 \\
Recomp-Abstractive & 2.84 & \textbf{0.71} & 3.55 \\
CompAct & 16.30 & 0.72 & 17.02 \\
\hline 

\met (BGE) & 0.08 & 0.82 & 0.90 \\
\met (Jina) & \textbf{0.04} & 0.82 & \textbf{0.86} \\
\met (GTE) & 0.12 & 0.82 & 0.94 \\
\hline
\end{tabular}

\caption{Compression latency, LLM read latency, and total latency per example averaged over 500 HQA examples using Gemma-2-9B-IT. Bold values denote the lowest metric in each column.}
\label{tab:efficiency-lat}
\end{table}

\begin{table}[h]
\centering
\footnotesize
\setlength{\tabcolsep}{6pt}
\renewcommand{\arraystretch}{1.2}

\begin{tabular}{l|c}
\hline
\textbf{Method}
& \textbf{Compression TFLOPs} \\
\hline

LLMLingua2
& 3.3 \\

DAC
& 35.2 \\

CPC
& 8.4 \\

LongLLMLingua
& 252.2 \\

Recomp$_{\mathrm{Extractive}}$
& 2.7 \\

CompAct
& 65.3 \\

\met (BGE)
& 2.1(Reranker) + 0.3(Sentence) = 2.5 \\

\met (Jina)
& 0.9(Reranker) + 0.3(Sentence) = \textbf{1.2} \\

\met (GTE)
& 1.1(Reranker) + 0.3(Sentence) = 1.4 \\

\hline
\end{tabular}

\caption{
Compression computational cost in TFLOPs averaged over 500 HQA examples with a batch size of 1. Lower values indicate lower computational cost.
}
\label{tab:efficiency-tflops}
\end{table}

\subsection{Memory Usage}

We evaluate the memory footprint of the compressor and the downstream LLM during inference. Memory usage is reported as the average peak GPU memory during the forward pass. For Recomp-Abstractive, per-example measurement is not supported, so we report the peak memory across 500 examples. Table~\ref{tab:memory_usage} reports the memory consumption of all methods. Our method achieves the second-lowest compressor memory usage among all frameworks while maintaining competitive memory consumption during downstream LLM inference.

\begin{table*}[h]
\centering
\footnotesize
\setlength{\tabcolsep}{10pt}
\renewcommand{\arraystretch}{1.2}

\begin{tabular}{l|c|c}
\hline
\textbf{Method} 
& \textbf{Compressor Memory (MB)} 
& \textbf{LLM Inference Memory (MB)} \\
\hline
Raw documents          & --         & 11,874.37 \\
LLMLingua2          & 2,588.44  &  7,158.68 \\
DAC          & 23,161.87  & 6,890.60  \\
CPC          & 2,969.61  & 6855.91  \\
LongLLMLingua          & 16,129.06  & 6,935.18  \\
Recomp-Extractive      & \textbf{945.46}     & 6,619.45  \\
Recomp-Abstractive     & 5,934.80   & 8,208.77  \\
Compact                & 14,090.37  & \textbf{6,580.15}  \\
\met (BGE)          & 1,252.01   & 7,202.59  \\
\met (Jina)         & 1,242.26   & 7,141.47  \\
\met (GTE)          & 1,323.48   & 7,178.40  \\
\hline
\end{tabular}

\caption{Peak GPU memory usage (in MB) for each compressor and the downstream LLM (Gemma-2-9B-IT). Memory is measured during the model forward pass. For Recomp-Abstractive, the peak across 500 examples is reported due to the lack of per-sample measurement.}
\label{tab:memory_usage}
\end{table*}

\subsection{Different Architectures for Sentence Selector}

We further evaluate our method with different encoder architectures to study whether compression performance depends heavily on the backbone model. In addition to ModernBERT (149M), we experiment with mGTE-base\footnote{https://huggingface.co/Alibaba-NLP/gte-multilingual-base} (305M) and mmBERT-mini\footnote{https://huggingface.co/jhu-clsp/mmBERT-small}~\citep{marone2025mmbertmodernmultilingualencoder} (140M). For each architecture, we train the model for three epochs and select the checkpoint with the best reward score for evaluation. We evaluate all models on the HQA and NQ benchmarks using the same training and evaluation setup. Table~\ref{tab:different-architecture} shows that all architectures achieve similar QA performance in terms of accuracy score, while producing substantially different compression ratios. Interestingly, the larger model does not consistently provide the best trade-off between compression and QA quality in our setting. For example, GTE achieves the highest compression ratio at 31.8x while maintaining performance comparable to ModernBERT. Meanwhile, the lightweight mmBERT-mini model also performs competitively in terms of $\mathrm{C}_{\beta}$ despite its smaller size. These results suggest that architecture choice mainly affects compression efficiency rather than downstream QA performance.

\begin{table*}[h]
\centering
\footnotesize
\setlength{\tabcolsep}{3pt}
\renewcommand{\arraystretch}{1.25}

\begin{tabular}{l|rccc|rccc}
\hline
\textbf{Method}
& \multicolumn{4}{c|}{\textbf{HQA}}
& \multicolumn{4}{c}{\textbf{NQ}} \\
\cline{2-9}
& $1/\tau$ & Acc & $\mathrm{C}_{0.5}$ & $\mathrm{C}_{1}$
& $1/\tau$ & Acc & $\mathrm{C}_{0.5}$ & $\mathrm{C}_{1}$\\
\hline
Raw Documents
& 1.0x & 34.6 & 0.00 & 0.00
& 1.0x & 30.1 & 0.00 & 0.00\\
\hline
\multicolumn{9}{l}{\textit{Reranker-only methods}} \\
BGE reranker & 3.0x & 35.2 & 0.39 & 0.46  & 3.0x & 29.6 & 0.33 & 0.41\\
Jina reranker & 3.0x & 34.6 & 0.38 & 0.46 & 3.0x & 30.6 & 0.34 & 0.42\\
GTE reranker & 3.0x & 34.7 & 0.38 & 0.46 & 3.0x & 30.6 & 0.34 & 0.42\\
\hline
\multicolumn{9}{l}{\textit{Compression frameworks}} \\
LLMLingua2 & 5.4x & 25.0 & 0.29 & 0.38 & 5.3x & 23.8 & 0.28 & 0.37 \\
DAC & 5.1x & 24.7 & 0.29 & 0.38 & 5.2x & 22.7 & 0.27 & 0.35\\
CPC & 5.6x & 30.5 & 0.35 & 0.45 & 5.2x & 28.0 & 0.32 & 0.42\\
LongLLMLingua & 5.1x & 31.8 & 0.36 & 0.46 & 5.1x & 25.7 & 0.30 & 0.39\\
Recomp-Extractive & 29.2x & 25.6 & 0.3 & 0.41 & 29.8x & 22.8 & 0.27 & 0.37\\
Recomp-Abstractive & \textbf{132.1x} & 27.9 & 0.33 & 0.44 & 48.3x & 27.6 & 0.32 & 0.43 \\
CompAct & 48.7x & 31.2 & 0.36 & 0.47 & 44.2x & 27.8 & 0.32 & 0.43\\
\hline
\multicolumn{9}{l}{\textit{Our work: \met (Modernbert-base)}} \\
No reranker & 26.8x & \textbf{37.4} & \textbf{0.43} & \textbf{0.54} & 23.6x & 31.7 & \textbf{0.37} & \textbf{0.48}\\
\hdashline
BGE reranker  & 39.5x & 36.3 & 0.42 & 0.53 & 36.4x & 29.5 & 0.34 & 0.45\\
Jina reranker & 38.5x & 35.8 & 0.41 & 0.52 & 42.0x & 30.0 & 0.35 & 0.46 \\
GTE reranker & 40.4x & 35.5 & 0.41 & 0.52 & 42.3x & 30.4 & 0.35 & 0.46\\
\hline
\multicolumn{9}{l}{\textit{Our work: \met (mmBert-mini)}} \\
No reranker & 11.0x & 37.1 & 0.42 & 0.53 & 17.9x & \textbf{32.1} & \textbf{0.37} & \textbf{0.48}\\
\hdashline
BGE reranker  & 27.8x & 36.0 & 0.41 & 0.52 & 22.3x & 29.6 & 0.34 & 0.45\\
Jina reranker & 27.5x & 35.6 & 0.41 & 0.52 & 25.2x & 30.6 & 0.35 & 0.46 \\
GTE reranker & 27.9x & 35.5 & 0.41 & 0.52 & 24.3x & 30.4 & 0.35 & 0.46\\
\hline
\multicolumn{9}{l}{\textit{Our work: \met (GTE-base)}} \\
No reranker & 31.8x & 36.7 & 0.42 & 0.53 & 27.8x & 29.2 & 0.34 & 0.45\\
\hdashline
BGE reranker  & 49.0x & 35.7 & 0.41 & 0.52 & 42.8x & 29.0 & 0.34 & 0.45\\
Jina reranker & 48.5x & 35.4 & 0.41 & 0.52 & 52.0x & 29.3 & 0.34 & 0.45\\
GTE reranker & 49.7x & 35.2 & 0.40 & 0.52 & \textbf{52.7x} & 28.8 & 0.34 & 0.45\\
\hline
\end{tabular}

\caption{
Results on two QA benchmarks for Gemma-2-9B-IT.
We report compression ratio ($1/\tau$), accuracy, and $\mathrm{C}_{\beta}$. Higher is better. Best values are in bold. 
}
\label{tab:different-architecture}
\end{table*}

\subsection{Effect of Improved Phase I Labels}
\label{subsec:improved-phase1-labels}

We further investigate whether the quality of the supervision used in Phase I can be improved by leveraging the gold supporting sentences provided in a multi-hop dataset. Instead of assigning positive labels based on whether a sentence contains the ground-truth answer, we assign a positive label when its character-level overlap with a gold supporting sentence exceeds 50\%. We denote this alternative intermediate finetuning strategy as Phase I*. We evaluate three training configurations on the  validation set: Phase I* only, Phase I* followed by Phase II, and our original Phase I followed by Phase II. The results are reported in Table~\ref{tab:improved-phase1-labels}.

\begin{table}[h]
\centering
\footnotesize
\setlength{\tabcolsep}{4pt}
\renewcommand{\arraystretch}{1.2}

\begin{tabular}{l|cccc}
\hline
\textbf{Method}
& \textbf{Acc}
& \textbf{$1-\tau$}
& $\mathbf{C}_{0.5}$
& $\mathbf{C}_{1}$ \\
\hline
Phase I*
& 82.3 & \textbf{86.4} (\%) & 0.83 & 0.84 \\

Phase I* + Phase II
& 88.6 & 83.3 (\%) & 0.87 & \textbf{0.86} \\

Phase I + Phase II
& \textbf{91.6} & 75.6 (\%) & \textbf{0.88} & 0.83 \\
\hline
\end{tabular}

\caption{
Comparison of different Phase I labeling strategies on the  validation set. Phase I* uses gold supporting sentences to construct the intermediate finetuning labels. We report accuracy (Acc), compression rate, and $\mathrm{C}_{\beta}$. Higher values are better, and best values are in bold.
}
\label{tab:improved-phase1-labels}
\end{table}

The results show that using supporting-sentence supervision in Phase I leads to more aggressive compression. Phase I* alone achieves the highest compression rate of 86.4\%, although its accuracy is substantially lower than the two configurations that include Phase II. Adding Phase II to Phase I* increases the accuracy from 82.3\% to 88.6\%, while maintaining a relatively high compression rate of 83.3\%. This demonstrates the benefit of Phase II in improving QA performance even when stronger supporting-sentence supervision is used during Phase I.
Comparing Phase I* + Phase II with our original Phase I + Phase II, the two labeling strategies exhibit different trade-offs. Phase I* + Phase II achieves a higher compression rate (83.3\% vs.\ 75.6\%) and a higher $\mathrm{C}_{1}$ score (0.86 vs.\ 0.83), whereas our original approach achieves higher QA accuracy (91.6\% vs.\ 88.6\%) and a slightly higher $\mathrm{C}_{0.5}$ score (0.88 vs.\ 0.87). These results suggest that supporting-sentence annotations can provide useful supervision when greater compression is desired, but do not necessarily improve QA accuracy.

Furthermore, Phase I* relies on gold supporting-sentence annotations, which may not be available in other QA datasets. In contrast, our original Phase I constructs labels using only the ground-truth answers, making the training procedure applicable to a broader range of QA datasets without requiring additional sentence-level annotations.

\subsection{Effect of Sentence Representation Tokens}
\label{subsec:cls-sen}

Our sentence selector uses a dedicated \texttt{[SEN]} token to represent each sentence and make sentence-level retention decisions. Although the standard \texttt{[CLS]} token can serve a similar role, we use \texttt{[SEN]} based on the intuition that \texttt{[CLS]} is generally used to represent the entire input sequence, whereas \texttt{[SEN]} provides an explicit representation for each individual sentence. To examine this design choice, we replace each \texttt{[SEN]} token with \texttt{[CLS]} and evaluate both variants on the  validation set. The results are reported in Table~\ref{tab:cls-sen}.

\begin{table}[t]
\centering
\footnotesize
\setlength{\tabcolsep}{5pt}
\renewcommand{\arraystretch}{1.2}

\begin{tabular}{l|cccc}
\hline
\textbf{Method}
& \textbf{Acc}
& \textbf{$1-\tau$}
& $\mathbf{C}_{0.5}$
& $\mathbf{C}_{1}$ \\
\hline
\texttt{[CLS]}
& \textbf{93.5} & 67.4 (\%) & 0.87 & 0.78 \\

\texttt{[SEN]}
& 91.6 & \textbf{75.6} (\%) & \textbf{0.88} & \textbf{0.83} \\
\hline
\end{tabular}

\caption{
Comparison between \texttt{[CLS]} and \texttt{[SEN]} tokens for sentence-level representation on the  validation set. We report accuracy (Acc), compression rate, and $\mathrm{C}_{\beta}$. Higher values are better, and best values are in bold.
}
\label{tab:cls-sen}
\end{table}

The results show that both tokens can effectively support sentence-level selection, but they exhibit different trade-offs between QA accuracy and compression. Using \texttt{[CLS]} achieves a higher accuracy of 93.5\%, compared with 91.6\% for \texttt{[SEN]}. However, \texttt{[SEN]} achieves a substantially higher compression rate of 75.6\%, compared with 67.4\% for \texttt{[CLS]}.
When considering both objectives jointly, \texttt{[SEN]} achieves higher $\mathrm{C}_{0.5}$ and $\mathrm{C}_{1}$ scores of 0.88 and 0.83, respectively, compared with 0.87 and 0.78 for \texttt{[CLS]}. These results indicate that although \texttt{[CLS]} favors QA accuracy, the dedicated \texttt{[SEN]} token provides a better overall balance between preserving QA performance and achieving effective compression.

\subsection{Additional Ablation Study on Different Finetuning Strategies}
\label{sec:more-ablation-study}
\begin{figure*}[h]
    \small
    \centering
    (A) 2WikiMultiHopQA\\
    \begin{minipage}{0.42\linewidth}
        \includegraphics[width=\linewidth]{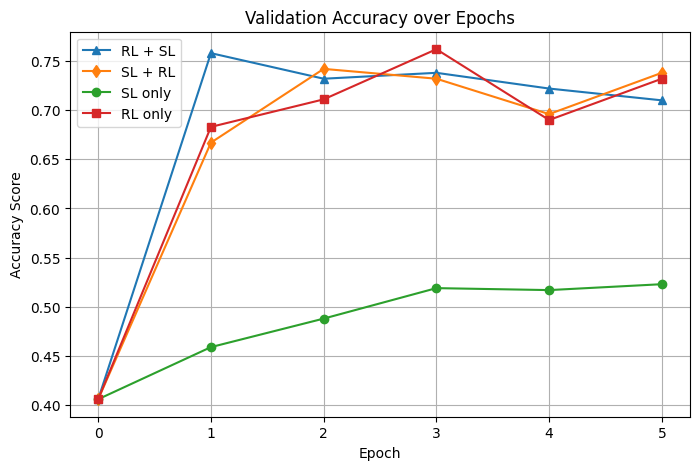}
    \end{minipage}
    \begin{minipage}{0.42\linewidth}
        \includegraphics[width=\linewidth]{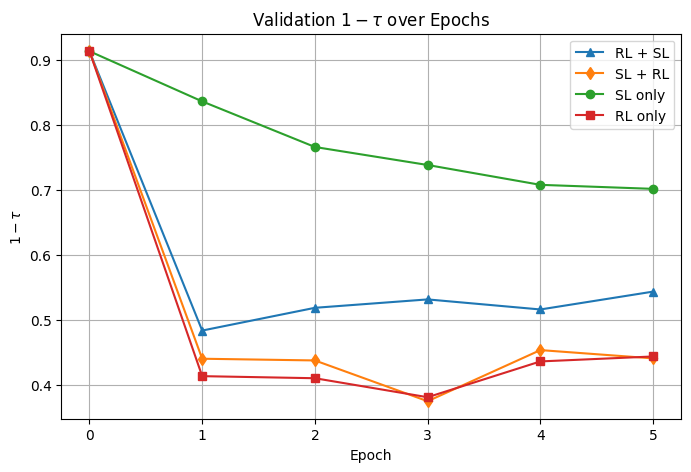}
    \end{minipage}
    \vspace{1em}

    (B) Natural Questions\\
    \begin{minipage}{0.42\linewidth}
        \includegraphics[width=\linewidth]{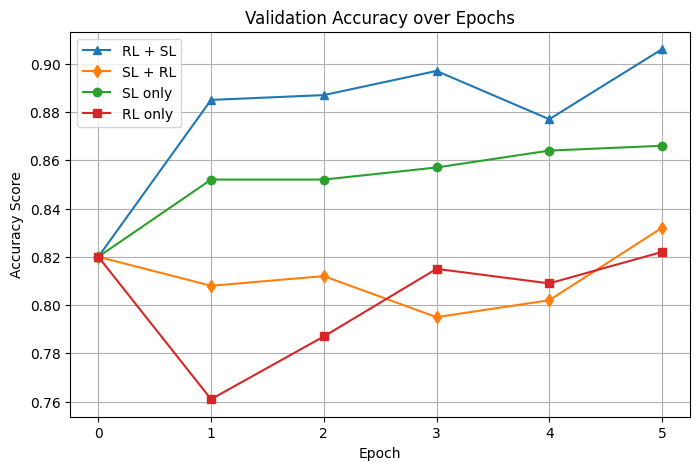}
    \end{minipage}
    \begin{minipage}{0.42\linewidth}
        \includegraphics[width=\linewidth]{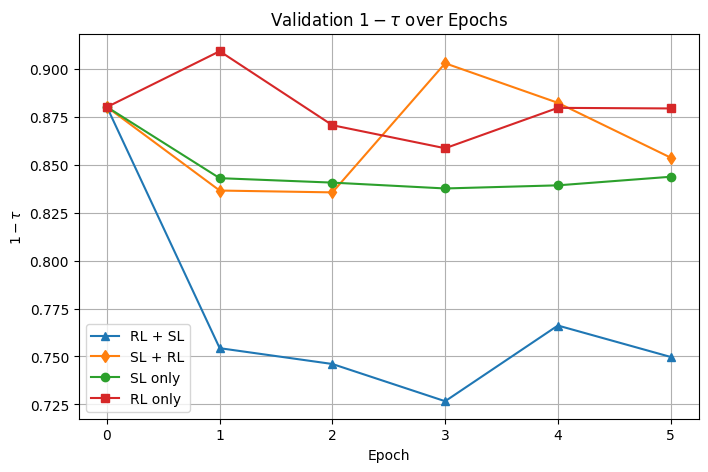}
    \end{minipage}
    \vspace{1em}

    (C) TriviaQA\\
    \begin{minipage}{0.42\linewidth}
        \includegraphics[width=\linewidth]{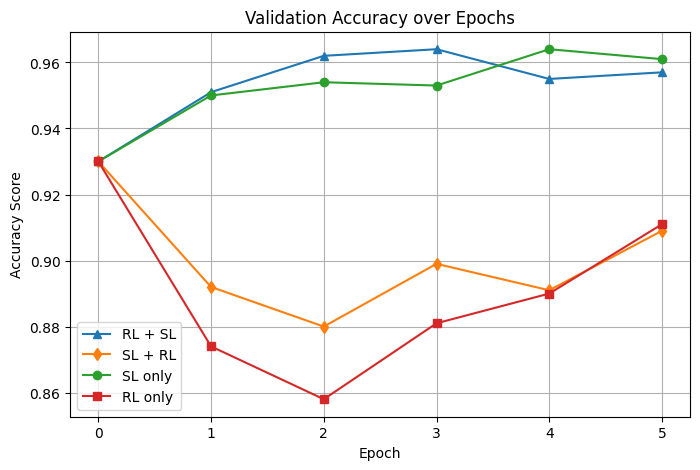}
    \end{minipage}
    \begin{minipage}{0.42\linewidth}
        \includegraphics[width=\linewidth]{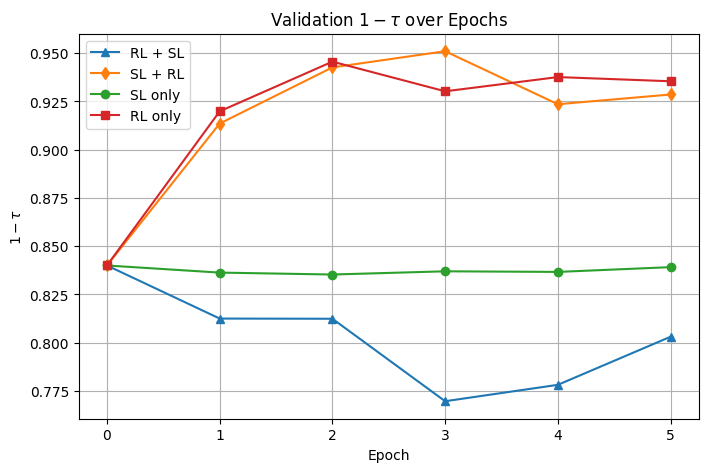}
    \end{minipage}
    
    \caption{Ablation study of different training strategies for \met\ on three QA datasets. 
    The plots compare (left) validation accuracy and (right) compression rate $1-\tau = 1 - |\mathbb{C}|/|\mathbb{D}|$ during validation across five epochs. For both metrics, the higher values are better.}
    \label{fig:additional-training-ablation}
\end{figure*}

In addition to RQ4 in Section~\ref{sec:analysis}, we conducted an ablation study on 2WikiQA, NQ, and TQA, evaluating four training strategies: reinforcement learning (RL) only, supervised learning (SL) only with the best trial memory, and the hybrid approach of \met that combines both approaches (RL+SL and SL+RL). 
The results are displayed in Figure~\ref{fig:additional-training-ablation}.
We observed similar patterns in HQA, NQ, and TQA, where intermediate finetuning led to higher accuracy.
Specifically, the hybrid strategy (RL+SL) often achieved higher validation accuracy in early epochs, while having a relatively lower compression rate.
In contrast, the RL only strategy prioritized the compression rate while sacrificing the accuracy. 
Meanwhile, the SL only strategy usually achieved higher validation accuracy than the RL one but with a lower and stable compression rate across epochs.
For 2WikiQA, where intermediate finetuning did not substantially improve accuracy, the trend of the hybrid (RL+SL) approach is similar to the other three datasets, whereas the trends of the RL only and SL only strategies are the opposite of those in the other three datasets.

\section{The Use of Large Language Models}
We leveraged ChatGPT and Gemini to enhance grammar, refine phrasing, and optimize word choice across the paper. We also used them to help format tables and figures in LaTeX. The typical prompts we used with the LLMs were concise instructions such as ``Proofread:'' or ``polish this text'', which direct the models to refine grammar and phrasing without altering the technical content.

\section{Licenses}

Our experiments use several publicly available datasets. HotpotQA is licensed under CC BY-SA 4.0. Natural Questions, 2WikiMultiHopQA, and TriviaQA are licensed under the Apache License 2.0. LongBench is licensed under the MIT License. All intermediate finetuning models used in our experiments, including Gemma-2-9B-IT and LLaMA-3.1-8B-Instruct, are used in accordance with their respective licenses for research purposes. 

\end{document}